\documentclass[runningheads]{llncs}

\usepackage{eccv}

\usepackage{eccvabbrv}
\usepackage{pdfpages}
\usepackage{graphicx}
\usepackage{booktabs}
\usepackage{graphicx}
\usepackage{amsmath}
\usepackage{amssymb}
\usepackage{booktabs}
\usepackage{float}
\usepackage{bm}
\usepackage{bbm}
\usepackage{mathtools}
\usepackage{amsfonts}       
\usepackage{nicefrac}       
\usepackage{microtype}      
\usepackage{multirow}
\usepackage{adjustbox}
\usepackage{wrapfig}
\usepackage{pifont}
\usepackage{placeins}
\usepackage{makecell}
\usepackage[normalem]{ulem}
\useunder{\uline}{\ul}{}

\newif\ifshowappendix
\showappendixtrue 
\usepackage[accsupp]{axessibility}  

\usepackage{hyperref}
\usepackage{marvosym}
\usepackage{orcidlink}

\newcommand{\rv}[1]{{{#1}}}

\begin{document}

\title{AdaCLIP: Adapting CLIP with Hybrid Learnable Prompts for Zero-Shot Anomaly Detection} 
\titlerunning{AdaCLIP}

\author{Yunkang Cao\inst{1,2}\orcidlink{0000-0001-7619-6618} \and
Jiangning Zhang\inst{3,4}\orcidlink{0000-0001-8891-6766} \and
Luca Frittoli\inst{2}\orcidlink{0000-0002-8205-4007} \and
Yuqi Cheng\inst{1} \orcidlink{0000-0003-1994-7301} \and \\
Weiming Shen\inst{1}\textsuperscript{\Letter}\orcidlink{0000-0001-5204-7992} \and
Giacomo Boracchi\inst{2}\orcidlink{0000-0002-1650-3054}
}

\authorrunning{Y.~Cao et al.}

\institute{Huazhong University of Science and Technology \\
\email{\{cyk\_hust,yuqicheng,shenwm\}@hust.edu.cn}\\ \and
Politecnico di Milano\\
\email{\{luca.frittoli,giacomo.boracchi\}@polimi.it} \and
Zhejiang University \and
Youtu Lab, Tencent \\
\email{186368@zju.edu.cn}
} 

\maketitle

\begin{abstract}
Zero-shot anomaly detection (ZSAD) targets the identification of anomalies within images from arbitrary novel categories. This study introduces AdaCLIP for the ZSAD task, leveraging a pre-trained vision-language model (VLM), CLIP. AdaCLIP incorporates learnable prompts into CLIP and optimizes them through training on auxiliary annotated anomaly detection data. Two types of learnable prompts are proposed: \textit{static} and \textit{dynamic}. Static prompts are shared across all images, serving to preliminarily adapt CLIP for ZSAD. In contrast, dynamic prompts are generated for each test image, providing CLIP with dynamic adaptation capabilities. The combination of static and dynamic prompts is referred to as hybrid prompts, and yields enhanced ZSAD performance. Extensive experiments conducted across 14 real-world anomaly detection datasets from industrial and medical domains indicate that AdaCLIP outperforms other ZSAD methods and can generalize better to different categories and even domains. Finally, our analysis highlights the importance of diverse auxiliary data and optimized prompts for enhanced generalization capacity. Code is available at \href{https://github.com/caoyunkang/AdaCLIP}{\url{https://github.com/caoyunkang/AdaCLIP}}.


\end{abstract}

\section{Introduction}
\label{sec:intro}

\begin{figure}[t]
  \centering
  \includegraphics[width=0.98\linewidth]{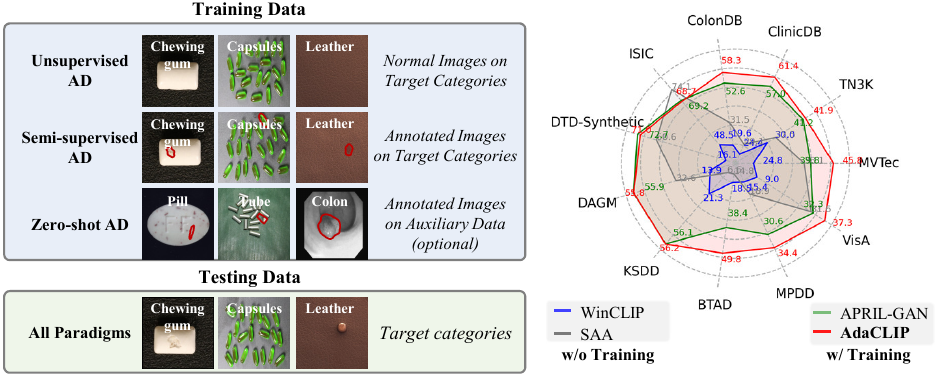}
  \caption{Left: Illustrations for training and test data of unsupervised, semi-supervised, and zero-shot anomaly detection paradigms. Right: Quantitative comparison with popular methods by pixel-level max-F1~\cite{WinClip} on industrial and medical datasets.}
  \label{fig:teaser}
\end{figure}

Anomaly detection (AD) in images~\cite{AD_SURVEY, cao2024survey} holds significant importance across various domains, including industrial inspection~\cite{MVTec-AD, Real3D, Real-iad} and medical diagnosis~\cite{ccalli2021deep}. The primary goal of AD methods is to detect deviations from normal patterns, either image or pixel-level. Most AD methods rely on unsupervised learning~\cite{Patchcore, CDO} and semi-supervised learning~\cite{ruffdeep,BiaS,ding2022catching} paradigms that require either normal samples or annotated anomalous samples from the target category for training, as depicted in Fig.~\ref{fig:teaser}. For instance, to train a dedicated model for the category `chewing gum', traditional unsupervised AD methods require a substantial dataset comprising normal `chewing gum' images, while semi-supervised approaches impose an even stricter requirement, requiring annotated abnormal images.

Some scenarios are characterized by the \textit{cold start} problem, meaning that it is not feasible to gather enough normal images for training an unsupervised model, thus preventing both unsupervised and semi-supervised AD solutions. The emerging zero-shot anomaly detection (ZSAD~\cite{WinClip}) paradigm addresses this issue, aiming at detecting anomalies in images belonging to unseen categories, without requiring any image of that category for training. Existing ZSAD methods commonly rely on pre-trained vision-language models (VLMs) due to their broad generalization capability. Some ZSAD methods employ VLMs for ZSAD without any additional training~\cite{Tamura_2023_BMVC,WinClip}, while others leverage annotated images from auxiliary anomaly-detection datasets to tailor VLMs for ZSAD, as Fig.~\ref{fig:teaser} shows.

The pioneering ZSAD method, WinCLIP~\cite{WinClip}, directly uses pre-trained VLMs with hand-crafted textual prompts to identify anomalies. Similarly to zero-shot classification, WinCLIP detects as anomalous images that are close to the selected prompts in the embedding space. However, WinCLIP exhibits limited detection performance since its underlying VLM, CLIP~\cite{clip}, is trained on natural image-text datasets~\cite{Laion400} and is not specialized for anomaly detection. Conversely, APRIL-GAN~\cite{APRIL-GAN} and AnomalyCLIP~\cite{AnomalyCLIP} address ZSAD by adapting VLMs on auxiliary anomaly-detection datasets that contain annotated anomalies. This adaptation scheme is gaining popularity due to the growing availability of annotated AD datasets~\cite{MVTec-AD,VisA} spanning diverse categories~\cite{MVTec-AD} and domains~\cite{VisA, tn3k}. Importantly, the adaptation scheme adheres to the zero-shot learning paradigm, as long as testing images do not belong to categories presented in the auxiliary AD dataset. 

The rationale behind ZSAD approaches is that testing images may exhibit universal patterns, either normal or anomalous, that VLMs can identify. 
Additionally, adapting VLMs on auxiliary data can be beneficial as these data might contain patterns that are useful for detecting anomalies in novel categories. For example, the scratches on `pill' images might improve the model's ability to detect similar abnormal patterns on `chewing gum' (as illustrated in Fig.~\ref{fig:teaser}). 


To take the most from auxiliary datasets for ZSAD, we propose AdaCLIP, which builds upon the mainstream zero-shot learning principle in CLIP. In particular, AdaCLIP computes similarities between patch embeddings and text embeddings for textual captions describing normal/abnormal states using CLIP. To enhance the ZSAD performance, AdaCLIP introduces additional lightweight learnable parameters in two forms: \emph{projection} and \emph{prompting} layers. \rv{As in APRIL-GAN~\cite{APRIL-GAN}, our projection layer is designed to align the dimensions between patch tokens and text embeddings, while introducing additional learnable parameters for fine-tuning CLIP.} Prompting layers are used to replace the original transformer layers within CLIP, by concatenating additional prompting tokens and the layer input. Prompting has proven very effective in adapting VLMs~\cite{MaPLe}. To ease the adaptation with auxiliary data, static and dynamic learnable prompts are introduced, where static prompts are shared across all images and dynamic prompts are generated based on the testing image. The combination of static and dynamic prompts, referred to as hybrid prompts, demonstrates significant generalization capabilities and promising ZSAD performance, as shown in Fig.~\ref{fig:teaser}. 

In summary, our contributions include the following key components:
\begin{itemize}
    \item We introduce a novel ZSAD method named AdaCLIP. AdaCLIP comprises hybrid (static and dynamic) learnable prompts to better exploit the auxiliary data to enhance ZSAD performance. A hybrid-semantic fusion module is also developed to extract region-level context about anomaly regions, thereby enhancing image-level anomaly detection performance. 
    \item We show that different VLMs --not only CLIP-- can be effectively adapted for ZSAD. Additionally, we demonstrate the importance of optimized prompts for detecting anomalies within individual images.
\end{itemize}

Our experiments demonstrate that we achieve state-of-the-art (SOTA) performance in ZSAD across 14 datasets spanning industrial and medical domains. We showcase that our AdaCLIP can effectively leverage information from auxiliary datasets, even when referring to categories from different domains (medical/industrial), outperforming alternative ZSAD methods. Additionally, we underscore that leveraging diverse auxiliary data is beneficial for ZSAD.

\section{Related Work}
\label{sec:related_work}

\subsection{Traditional Anomaly Detection}

\noindent\textbf{Unsupervised Anomaly Detection} methods like~\cite{IKD,Patchcore} learn exclusively from normal samples within target categories. Unsupervised AD methods typically model normal sample distributions during training and subsequently compare test samples to the learned normal sample distribution to detect anomalies. A very effective approach consists in extracting features from each sample using pre-trained neural networks~\cite{RD4AD,2023CaiDis,CDO}, and then modeling the features distribution by knowledge distillation~\cite{2023liuST,jiang_masked_2023,vitad}, reconstruction~\cite{DSR,diad,bionda2022deep,yao_feature_2022}, or memory bank-based approaches~\cite{GCPF,regad}. 

\noindent\textbf{Semi-supervised Anomaly Detection} methods like~\cite{BiaS,ding2022catching} require both normal and abnormal images with annotations from target categories for training. They typically utilize annotated abnormal samples to learn a more compact description boundary for normal samples. Since some additional abnormal samples are exploited, they typically present better AD performance in comparison to unsupervised AD but impose a strict requirement for data.

Despite the promising anomaly detection performance achieved by these traditional AD methods, their effectiveness tends to diminish when fewer normal samples are available for training. In contrast, we aim to develop a generic ZSAD model for anomaly detection across unseen categories without training samlpes.

\subsection{Zero-shot Anomaly Detection}
Zero-shot learning often requires extensive training data to attain generalization abilities~\cite{WinClip,clipad}. Many off-the-shelf VLMs have been developed, presenting promising zero-shot capabilities. These pre-trained VLMs are leveraged to identify anomalies across unbounded categories. For instance, WinCLIP~\cite{WinClip} employs CLIP~\cite{clip} to compute similarities between embeddings of image patches and embeddings of captions regarding normal/abnormal states, which is subsequently enhanced by text augmentation in~\cite{Tamura_2023_BMVC}. In contrast, SAA~\cite{SAA} utilizes Grounding DINO~\cite{Grounding-dino} to identify abnormal regions within a test image using text prompts, followed by refinement with SAM~\cite{SAM}. However, these VLMs are typically trained on natural image-text pairs and are not specifically designed for AD. Therefore, APRIL-GAN~\cite{APRIL-GAN} and CLIP-AD~\cite{clipad} enhance the ZSAD performance of CLIP by tuning additional projection layers with annotated auxiliary AD data. With these auxiliary data, AnomalyCLIP~\cite{AnomalyCLIP} preliminarily explores prompt learning and introduces learnable text prompts to adapt VLMs for ZSAD. AnomalyGPT~\cite{AnomalyGPT} also introduces textual prompting learning but for unsupervised AD. In this paper, we further delve into prompt learning and develop multimodal hybrid learnable prompts to maximize the utility of auxiliary AD data. Table 4 in Appendix highlights the significance of AdaCLIP in comparison to other alternatives.

\subsection{Prompt Learning}

In the realm of VLMs, prompt learning~\cite{MaPLe} involves incorporating learnable tokens into the input image or text, effectively tailoring VLMs to specific scenarios. 
Early prompt learning methods introduce static prompts to VLMs. For instance, CoOp~\cite{CoOp} integrates learnable tokens in addition to the input text into the text branch. However, recent advancements in prompt learning methods~\cite{zha2023_IDPT} have identified that static prompts may be susceptible to distribution diversity. Consequently, CoCoOp~\cite{coCoOp} and IDPT~\cite{zha2023_IDPT} propose generating dynamic prompts based on the inputs for improving modeling capabilities.  \rv{
Whereas previous prompt learning methods primarily focused on the text encoder of VLMs~\cite{coCoOp}, recent studies~\cite{MaPLe,VPT} have increasingly acknowledged the significance of prompting the image encoder, \ie, visual prompting, to better exploit the multimodal capabilities of VLMs. In this paper, we propose multimodal (image+text) hybrid (static+dynamic) prompts to adapt VLMs for improving anomaly detection.  
}

\section{Problem Formulation}

\rv{Our objective is to develop a model that associates an input image $\mathbf{I} \in \mathbb{R}^{H \times W \times 3}$ with an image-level anomaly score $S$ and a pixel-level anomaly map $\mathbf{M} \in \mathbb{R}^{H \times W}$, indicating whether $\mathbf{I}$ and its pixels are normal or abnormal. Typically, the values of the anomaly score and anomaly map pixels fall within the range $[0, 1]$, where larger values indicate higher probabilities of being abnormal.} \rv{We operate within the ZSAD context, training our model using an auxiliary anomaly detection dataset 
$\mathcal{I}_{\text{train}} = \{(\mathbf{I}_i, \mathbf{G}_i)\}_{i=1}^{N_{\text{train}}}$, which contains categories distinct from those in the testing dataset $\mathcal{I}_{\text{test}} = \{\mathbf{I}_i\}_{i=1}^{N_{\text{test}}}$. The auxiliary training dataset includes both normal and abnormal images $\mathbf{I}$ along with their annotated masks $\mathbf{G}\in \mathbb{R}^{H \times W}$, where pixels have value 0 if normal and 1 if abnormal. 
By learning from this auxiliary dataset $\mathcal{I}_{\text{train}}$, the model is expected to learn normal and abnormal patterns that are common to different classes, enabling the detection of anomalies in novel categories.
}
 
\begin{figure}[t]
  \centering
  \includegraphics[width=0.98\linewidth]{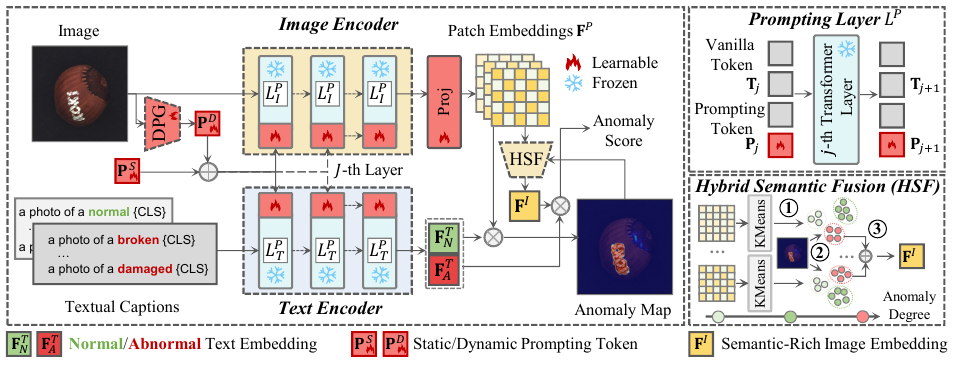}
  \caption{\textbf{Framework of AdaCLIP.} 
  }
  \label{fig:framework}
\end{figure}

\section{AdaCLIP}\label{sec:method}

\subsection{Overview}
The framework of AdaCLIP is illustrated in Fig.~\ref{fig:framework}. Given an image $\mathbf{I}$, AdaCLIP follows the general ZSAD principle of comparing CLIP embedding as WinCLIP~\cite{WinClip} do. In particular, we detect anomalies by calculating similarities in CLIP embedding space between the image and textual captions for normal/abnormal states, such as \texttt{"A photo of normal [CLS]"} and \texttt{"A photo of damaged [CLS]"}, where \texttt{[CLS]} denotes to the name of the testing category, like `carpet', `hazelnut', \etc.  
Notably, AdaCLIP enhances the pre-trained CLIP by incorporating learnable parameters through prompting layers for image and text encoders, denoted as $L_I^P$ and $L_T^P$ respectively, which replace the original transformer layers. For the prompting layers, both static prompts $\mathbf{P}^S$ and dynamic prompts $\mathbf{P}^D$ are introduced. AdaCLIP also introduces a projection layer $\text{Proj}$ at the end of the image encoder, and a Hybrid Semantic Fusion (HSF) module designed to extract semantic-rich image embeddings for computing image-level anomaly scores $S$. 

\subsection{Prompting Layers}
AdaCLIP introduces prompting layers $L_I^P$ and $L_T^P$ to replace the original transformer layers in the image and text encoders of CLIP, respectively. Prompting layers~\cite{MaPLe} preserves the weights of the transformer (to inherit its generalization ability) but concatenates learnable prompting tokens $\mathbf{P}$ to \rv{the vanilla tokens derived from the input images or texts}, as illustrated in Fig.~\ref{fig:framework}. Thanks to the self-attention mechanism in transformer layers, the learnable prompting token will contribute to all the output tokens, including the vanilla ones. 

More specifically, prompting tokens $\mathbf{P} \in \mathbb{R}^{M \times C}$ are concatenated to the input vanilla tokens $\mathbf{T} \in \mathbb{R}^{N \times C}$ of the transformer layer. Here, $C$ denotes the embedding dimension, while $N$ and $M$ denote the lengths of vanilla tokens and prompting tokens, respectively, where $M \ll N$ for lightweight adaptation. Let $L^P_j$ denote the $j$-th prompting layer, then the feed-forward process is,
\begin{align}
    [\mathbf{T}_{j+1}, \_] = L^P_j([\mathbf{T}_j, \mathbf{P}_j]),  \quad j \leq J, \\
    [\mathbf{T}_{j+1}, \mathbf{P}_{j+1}] = L^P_j([\mathbf{T}_j, \mathbf{P}_j]) , \quad j > J, 
\end{align}

\noindent \rv{where \([\cdot,\cdot]\) denotes concatenation along rows}. Learnable prompting tokens are incorporated up to a limited depth $J$, while prompting tokens for the remaining layers are generated through feed-forwarding. Typically, $J$ is set to a small value, as too many learnable parameters may result in overfitting on auxiliary data.

\subsection{Hybrid Learnable Prompts}
To effectively utilize auxiliary data for enhanced anomaly detection performance, we introduce both \textit{static} and \textit{dynamic} prompts.



\noindent\textbf{Static Prompts $\mathbf{P}^S$.} Static prompts $\mathbf{P}^S$ serve as foundational learning tokens shared across all images, which are explicitly learned from auxiliary data during training, as Fig~\ref{fig:framework} shows.  However, their limited adaptation effectiveness is acknowledged by previous studies~\cite{zha2023_IDPT}. 


\noindent\textbf{Dynamic Prompts $\mathbf{P}^D$.} We further introduce dynamic prompts $\mathbf{P}^D$ to enhance the modeling capacity for diverse distributions. Dynamic prompts differ from static prompts as they are generated on each testing image by the Dynamic Prompt Generator (DPG):
\begin{equation}
\mathbf{P}^D = \text{DPG}(\mathbf{I}).
\end{equation}
\noindent In our case DPG is a frozen pre-trained backbone such as CLIP to extract class tokens, followed by a learnable linear layer to project the class tokens into dynamic prompts $\mathbf{P}^D$. Both dynamic prompts for $L^P_I$ in the image encoder and $L^P_T$ in the text encoder are generated from the testing image, as shown in Fig.~\ref{fig:framework}.

AdaCLIP sums up the static and dynamic prompts, referred to as hybrid prompts, for both prompting layers $L^P_I$ and $L^P_T$. By replacing the original transformer layers with these prompting layers, the image encoder extracts patch embeddings $\mathbf{F}^P=\{\mathbf{F}^P_0, \ldots\}$ for the input image $\mathbf{I}$ from multiple prompting layers, while the text encoder generates normal/abnormal text embeddings $\mathbf{F}^T_N$, $\mathbf{F}^T_A$ for the corresponding textual captions.

\subsection{Projection Layer}
The original CLIP~\cite{clip} architecture makes the dimensions of patch embeddings and text embeddings unmatched, thus we append a projection layer {Proj} to the image encoder. In particular, we align the dimensions between patch embeddings ($\mathbf{F}^P$) and the embeddings of normal ($\mathbf{F}^T_N$) and anomalous ($\mathbf{F}^T_A$) texts by introducing a linear layer with bias. In addition, the projection layer adds some learnable parameters for CLIP adaption.

\subsection{Pixel-Level Anomaly Localization}
We derive the anomaly score by measuring the cosine similarities between patch embeddings $\mathbf{F}^P$, and text embeddings $\mathbf{F}^T_N$ and $\mathbf{F}^T_A$. We adopt the same approach as in WinCLIP~\cite{WinClip}, and define the anomaly map from $i$-th layer as follows: 
\begin{equation}~\label{eq:AL}
\mathbf{M}_i = \phi\left(\frac{\exp(\cos(\mathbf{F}^P_i, \mathbf{F}^T_A))}{\exp(\cos(\mathbf{F}^P_i, \mathbf{F}^T_N)) + \exp(\cos(\mathbf{F}^P_i, \mathbf{F}^T_A))}\right), 
\end{equation}
\noindent where $\cos(\cdot, \cdot)$ denotes the cosine similarity and $\phi$ is a reshape and interpolate function, transforming anomaly scores for patch embeddings into anomaly maps $\mathbf{M}_i \in \mathbb{R}^{H \times W}$. Then we take anomaly maps from several layers in a multi-hierarchy manner~\cite{WinClip} and aggregate these anomaly maps into a final prediction $\mathbf{M}$. During training, AdaCLIP optimizes the pixel-level anomaly map $\mathbf{M}$ with dice loss~\cite{diceloss} and focal loss~\cite{focalloss} on the auxiliary data.

\subsection{Hybrid Semantic Fusion Module}
AdaCLIP introduces an HSF module to improve image-level AD performance. \rv{Traditional AD methods~\cite{CDO, APRIL-GAN} for image-level AD often select the maximum values of anomaly maps as anomaly scores, but this is sensitive to noisy predictions. In contrast, we present the HSF module to aggregate patch embeddings that are more likely to represent abnormalities, thereby aggregating region-level information for robust image-level anomaly detection.} We refer to the output of HSF as semantic-rich embedding $\mathbf{F}^I$.


As Fig.~\ref{fig:framework} shows, the HSF module follows a three-step paradigm: 
\ding{172} Cluster patch embeddings into $K$ groups using KMeans~\cite{k-means++}.
\ding{173} Compute the anomaly scores of individual clusters by averaging the scores of the corresponding positions in the anomaly map $\mathbf{M}$.
\ding{174} Select the cluster with the highest anomaly scores, calculate its centroids, and aggregate them into the final semantic-rich image embedding $\mathbf{F}^I$, which encapsulates semantic information about the most abnormal clusters. 
The resulting semantic-rich image embedding effectively improves image-level AD performance compared to the maximum value-based anomaly detection methods. More details regarding HSF are presented in Appendix Section 2.2.


\subsection{Image-Level Anomaly Detection}
\rv{After extracting the semantic-rich image embeddings $\mathbf{F}^I$, we compute the image-level anomaly scores $S$ similar to~\eqref{eq:AL}, using cosine similarities between $\mathbf{F}^I$ and the text embeddings $\mathbf{F}_A^N$ and $\mathbf{F}_A^T$, followed by softmax normalization.
} Then we optimize image-level anomaly scores $S$ using focal loss~\cite{focalloss}. 

\section{Experiments}
\label{sec:exp}

\subsection{Experimental Setup}\label{sec:experimental_setup}

\noindent\textbf{Datasets.}
We conduct experiments using datasets from industrial and medical domains. Specifically, for the industrial domain, we use MVTec AD~\cite{MVTec-AD}, VisA~\cite{VisA}, MPDD~\cite{MPDD}, BTAD~\cite{BTAD}, KSDD~\cite{SDD}, DAGM~\cite{DAGM}, and DTD-Synthetic~\cite{DTD} datasets. In the medical domain, we consider brain tumor detection datasets HeadCT~\cite{HeadCT}, BrainMRI~\cite{BrainMRI}, Br35H~\cite{Br35h}, skin cancer detection dataset ISIC~\cite{ISIC}, colon polyp detection datasets ClinicDB~\cite{ClinicDB}, and ColonDB~\cite{ColonDB}, as well as thyroid nodule detection dataset TN3K~\cite{tn3k}. A detailed introduction to these datasets can be found in Appendix Section 1. 

\noindent\textbf{Evaluation Metrics.} 
Following previous ZSAD studies~\cite{WinClip, APRIL-GAN}, we employ the Area Under the Receiver Operating Characteristic Curve (AUROC) and the maximum F1 score (max-F1) under the optimal threshold to evaluate both image-level and pixel-level AD performance. In addition to dataset-level results, we also report domain-level average performance in the form of (AUROC, max-F1).


\noindent\textbf{Implementation Details.} 
This study employs the pre-trained CLIP (ViT-L/14@336px)\footnote{https://github.com/mlfoundations/open\_clip} as the default backbone and extracts patch embeddings from the 6-th, 12-th, 18-th, and 24-th layers. All images undergo resizing to a resolution of $518\times518$ for both training and testing. 
For the ZSAD task, it is imperative that the auxiliary data does not contain any categories present in the test set. Although ClinicDB~\cite{ClinicDB} and ColonDB~\cite{ColonDB} both comprise colon polyp data, their appearances differ significantly. Therefore, we default to using the industrial dataset, MVTec AD~\cite{MVTec-AD}, and the medical dataset, ClinicDB~\cite{ClinicDB}, as auxiliary data. For evaluations on MVTec AD and ClinicDB, VisA~\cite{VisA} and ColonDB~\cite{ColonDB} are utilized for training.
The prompting depth $J$ is set to four and the prompting length $M$ is set to five by default. We train AdaCLIP for five epochs with a learning rate of 0.01. All experiments are conducted using PyTorch-1.9.2 with a single NVIDIA A6000 48GB GPU. Appendix Section 3 presents further implementation details.

\subsection{Main Experimental Results}

\noindent\textbf{Comparison Methods.}
This study compares the proposed AdaCLIP with two sets of methods: with and without training on auxiliary data. For methods without training, we reproduce SAA~\cite{SAA} and WinCLIP~\cite{WinClip} for comparisons. Regarding methods with training, we choose the existing ZSAD method based on CLIP, APRIL-GAN~\cite{APRIL-GAN}, and AnomalyCLIP~\cite{AnomalyCLIP}. In addition, to explore whether other VLMs excluding CLIP can be adapted for ZSAD, we train DINOV2~\cite{dinov2} and SAM~\cite{SAM} on the auxiliary data by adding linear layers as segmentation heads after multiple transformer layers. More details about the implementation of these methods can be found in Appendix Section 3. \rv{Unfortunately, we cannot directly compare with AnomalyCLIP~\cite{AnomalyCLIP} because its implementation is not publicly available before our submission date. To enable a fair comparison, we have evaluated AdaCLIP under the experimental setting of AnomalyCLIP, and the results are reported in Appendix Section 4.} 




\noindent\textbf{Zero-shot Anomaly Detection in the Industrial Domain:}
Table~\ref{tab:industrial} reports the results in the industrial domain. It distinctly illustrates that methods with training exhibit superior performance compared to alternative ZSAD methods without training on auxiliary data. In particular, WinCLIP and SAA which utilize hand-crafted textual prompts present subpar AD performance. Conversely, adapting DINOV2 and SAM with auxiliary data demonstrates promising pixel-level ZSAD performance. 
The superior performance of the set of ZSAD methods trained with the auxiliary data underscores that pre-trained VLMs are already endowed with essential knowledge for anomaly detection. This existing knowledge can be effectively leveraged for ZSAD through proper adaptation, like the strategy we employed.

\begin{figure}[th!]
  \centering
  \includegraphics[width=0.98\linewidth]{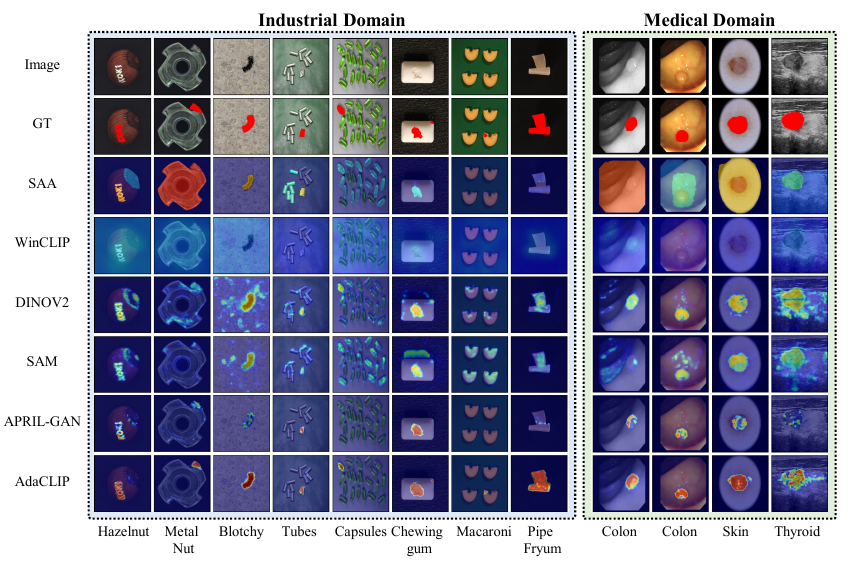}
  \caption{\textbf{Visualization of anomaly maps of different ZSAD methods.} The proposed AdaCLIP can get the most precise segmentation results for novel categories in both industrial and medical domains.}
  \label{fig:qualitative}
\end{figure}

Moreover, as evident in Table~\ref{tab:industrial}, the proposed AdaCLIP showcases significant improvements over other ZSAD methods, \eg, 3.7\% image-level and 3.3\% pixel-level enhancements on max-F1 compared to the second-place method. Also, AdaCLIP achieves the best overall ranking across all datasets in terms of both image- and pixel-level performance. This showcases the excellence of AdaCLIP and validates the efficacy of the introduced prompting layers. We further present visualizations of the predicted anomaly maps across various datasets in Fig.~\ref{fig:qualitative}. AdaCLIP exhibits significantly more accurate segmentation for novel industrial categories in comparison to other methods. The precise detection results for challenging categories such as tubes, capsules, and pipe fryum further highlight the superiority of AdaCLIP.

\begin{table*}[t!]
\centering
\caption{\textbf{Comparisons of ZSAD methods in the industrial domain.} The best performance is in \textbf{bold}, and the second-best is \underline{underlined}. $^\dag$ denotes to results taken from original papers. Rank denotes to the average performance rankings of different methods on various datasets.}
\label{tab:industrial}
\resizebox{1\linewidth}{!}{
\begin{tabular}{@{}cccccccc@{}}
\toprule[1.5pt]
\multirow{2}{*}{Metric} &
  \multirow{2}{*}{Dataset} &
  \multicolumn{2}{c}{w/o supervised training} &
  \multicolumn{4}{c}{w/ supervised   training} \\ \cmidrule(l){3-4} \cmidrule(l){5-8} 
 &
   &
 {SAA~\cite{SAA}} &
  {WinCLIP~\cite{WinClip}} &
  {DINOV2~\cite{dinov2}} &
  {SAM~\cite{SAM}} &
  {APRIL-GAN~\cite{APRIL-GAN}} &
  {\textbf{AdaCLIP}} \\ \midrule
\multirow{8}{*}{\begin{tabular}[c]{@{}c@{}}Image-level\\      (AUROC, max-F1)\end{tabular}} &
  MVTec AD &
  (63.5, 87.4) &
    (\textbf{91.8},
  \textbf{92.9})$^{\dag}$
  &
  (74.4,
  87.4) &
  (70.8,
  86.0) &
  (82.3, 
  88.9) &
  (\underline{89.2},
  \underline{90.6}) \\
 &
  VisA &
  (67.1,
  75.9) &
  (78.1, 
  \underline{80.7})$^{\dag}$ 
  &
  (75.2,
  78.5) &
  (61.9,
  73.9) &
  (\underline{81.7},
  {80.7}) &
  (\textbf{85.8}, 
  \textbf{83.1}) \\
 &
  MPDD &
  (42.7,
  73.9) &
  (61.4,
  \underline{77.5}) &
  (62.4,
  74.9) &
  (63.0,
  77.0) &
  (\underline{66.0},
  76.0) &
  (\textbf{76.0},
  \textbf{82.5}) \\
 &
  BTAD &
  (59.0,
  \textbf{89.7}) &
  (68.2,
  67.6) &
  (79.3,
  69.3) &
  (\textbf{89.4},
  85.7) &
  (85.2,
  82.0) &
  (\underline{88.6},
  \underline{88.2}) \\
 &
  KSDD &
  (68.6,
  37.6) &
  (93.3,
  79.0) &
  (94.9,
  77.5) &
  (65.8,
  37.9) &
  (\underline{95.7},
  \underline{85.2}) &
  (\textbf{97.1},
  \textbf{90.7}) \\
 &
  DAGM &
  (87.1,
  88.8) &
  (91.7,
  87.6) &
  (90.7,
  89.2) &
  (82.7,
  83.6) &
  (\underline{93.5},
  \underline{91.8}) &
  (\textbf{99.1},
  \textbf{97.5}) \\
 &
  DTD-Synthetic &
  (94.4,
  93.5) &
  (95.1,
  94.1) &
  (85.8,
  93.5) &
  (81.9,
  91.1) &
  (\textbf{98.1},
  \textbf{96.8}) &
  (\underline{95.5},
  \underline{94.7})\\ \cmidrule{2-8}
 &
  Average &
  (68.9,
  78.1) &
  (82.8,
  82.8) &
  (80.4,
  81.5) &
  (73.6,
  76.4) &
  (\underline{86.1},
  \underline{85.9}) &
 (\textbf{90.2},
  \textbf{89.6}) \\   & Rank  
 & (5.3, 4.4)
 & (3.4, 3.4)
 & (4.0, 4.1)   
 & (4.7, 5.0)
 & (\underline{2.1}, \underline{2.6})
 & (\textbf{1.4}, \textbf{1.4}) \\ \midrule
\multirow{8}{*}{\begin{tabular}[c]{@{}c@{}}Pixel-level\\      (AUROC, max-F1)\end{tabular}} &
  MVTec AD &
  (75.5,
  38.1) &
  (85.1,
  31.6)$^{\dag}$ &
 (\underline{85.9},
  39.6) &
  (85.4,
  29.4) &
  (83.7,
  \underline{39.8}) &
  (\textbf{88.7},
  \textbf{43.4}) \\
 &
  VisA &
  (76.5,
  31.6) &
  (79.6,
  14.8)$^\dag$ &
  (95.0,
  30.3) &
  (92.6,
  18.2) &
  (\underline{95.2},
  \underline{32.3}) &
  (\textbf{95.5},
  \textbf{37.7}) \\
 &
  MPDD &
 (81.7,
  18.9) &
  (71.2,
  15.4) &
  (\underline{95.6},
  \underline{31.1}) &
  (94.8,
  22.1) &
  (95.1,
  30.6) &
  (\textbf{96.1},
  \textbf{34.9}) \\
 &
  BTAD &
  (65.8,
  14.8) &
  (72.6,
  18.5) &
  (91.9,
  43.4) &
  (\textbf{93.8},
  \underline{46.9}) &
  (89.5,
  38.4) &
  (\underline{92.1},
  \textbf{51.7}) \\
 &
  KSDD &
  (78.8,
  6.6) &
  (95.8,
  21.3) &
  (\textbf{99.3},
  50.6)&
  (91.2,
  18.4) &
  (\underline{98.2},
  \textbf{56.2}) &
  (97.7,
  \underline{54.5}) \\
 &
  DAGM &
  (62.7,
  32.6) &
  (81.3,
  13.9) &
  (\underline{90.9},
  52.0) &
  (88.6,
  40.7) &
  (90.3,
  \textbf{57.9}) &
  (\textbf{91.5},
  \underline{57.5}) \\
 &
  DTD-Synthetic &
  (76.7,
  60.6) &
  (79.5,
  16.1) &
  (97.0,
  63.4) &
  (95.0,
  56.7) &
  (\underline{97.8},
  \textbf{72.7}) &
  (\textbf{97.9},
  \underline{71.6}) \\ \cmidrule{2-8}
 &
  {Average} &
  (73.9,
  29.0) &
  (80.7,
  18.8) &
  (\underline{93.7},
  44.3) &
  (91.7,
  33.2) &
 (92.8,
  \underline{46.9}) &
  (\textbf{94.2},
  \textbf{50.2}) \\ 
    & Rank  
 & (5.9, 4.7)
 & (4.9, 5.6)
 & (\underline{2.3}, 3.0)   
 & (3.6, 4.3)
 & (3.0, \underline{2.0})
 & (\textbf{1.4}, \textbf{1.4}) \\   \bottomrule[1.5pt]
\end{tabular}
}
\end{table*}
\begin{table}[t!]
\centering
\caption{\textbf{Comparisons of ZSAD methods in the medical domain.} 
The best performance is in \textbf{bold}, and the second-best is \underline{underlined}.  Rank denotes to the average performance rankings of different methods on various datasets.
}
\label{tab:medical}
\resizebox{1\linewidth}{!}{
\begin{tabular}{@{}cccccccc@{}}
\toprule[1.5pt]
\multirow{2}{*}{Metric} &
  \multirow{2}{*}{Dataset} &
  \multicolumn{2}{c}{w/o supervised training} &
  \multicolumn{4}{c}{w/ supervised   training} \\ \cmidrule(l){3-4} \cmidrule(l){5-8} 
 &
   &
 {SAA~\cite{SAA}} &
  {WinCLIP~\cite{WinClip}} &
  {DINOV2~\cite{dinov2}} &
  {SAM~\cite{SAM}} &
  {APRIL-GAN~\cite{APRIL-GAN}} &
  {\textbf{AdaCLIP}} \\ \midrule
\multirow{4}{*}{\begin{tabular}[c]{@{}c@{}}Image-level\\      (AUROC, max-F1)\end{tabular}} &
  HeadCT &
  (46.8,
  68.0) &
  (84.1,
  79.8) &
  (71.4,
  72.4) &
  (78.4,
  76.4) &
  (\textbf{93.6},
  \textbf{86.4}) &
  (\underline{91.4},
  \underline{85.2}) \\
 & BrainMRI 
 & (34.4, 76.7) 
 & (\underline{89.8}, 86.3) 
 & (78.3, 82.7)      
 & (71.5, 78.9) 
 & (89.7, \underline{89.5}) 
 & (\textbf{94.8}, \textbf{91.2}) \\
 & Br35H    
 & (33.2, 67.3)
 & (81.6, 74.4) 
 & (69.1, 70.5)     
 & (59.0, 67.2)
 & (\underline{95.6}, \underline{91.0})
 & (\textbf{97.7}, \textbf{92.4})\\ \cmidrule{2-8}
 & Average  
 & (38.1, 70.7)
 & (85.2, 80.2)
 & (72.9, 75.2)   
 & (69.7, 74.1)
 & (\underline{93.0}, \underline{89.0})
 & (\textbf{94.6}, \textbf{89.6}) \\
  & Rank  
 & (6.0, 5.7)
 & (2.7, 3.0)
 & (4.3, 4.3)   
 & (4.7, 5.0)
 & (\underline{2.0}, \underline{1.7})
 & (\textbf{1.3}, \textbf{1.3}) \\ \midrule
\multirow{5}{*}{\begin{tabular}[c]{@{}c@{}}Pixel-level\\      (AUROC, max-F1)\end{tabular}} &
  ISIC &
  (83.8,
  74.2) &
  (67.1,
  48.5) &
  (\textbf{94.2},
  \underline{79.6}) &
  (\underline{94.2},
  \textbf{81.0}) &
  (92.1,
  77.4) &
  (89.3,
  71.4) \\
 & ColonDB  
 & (71.8, 31.5)
 & (61.1, 19.6)
 & (87.3, \underline{56.5})
 & (86.1, 45.7)
 & (\underline{88.7}, 52.6)   
 & (\textbf{90.4}, \textbf{58.2}) \\
 & ClinicDB 
 & (66.2, 29.1) 
 & (67.1, 24.4)
 & (83.3, \underline{56.2})
 & (\underline{83.5}, 43.0)
 & (82.5, 51.8)      
 & (\textbf{84.4}, \textbf{58.2}) \\
 & TN3K     
 & (66.8, 32.6)
 & (67.2, 30.0)
 & (73.3, 35.7)     
 & (70.1, 32.5)
 & (\underline{75.9}, \underline{36.4})
 & (\textbf{77.2}, \textbf{41.9}) \\ \cmidrule{2-8}
 & Average  
 & (72.1, 41.8)
 & (65.6, 30.6)
 & (84.5, \underline{57.0})
 & (83.5, 50.5)
 & (\underline{84.8}, 54.6)     
 & (\textbf{85.3}, \textbf{57.4}) \\
  & Rank  
 & (5.5, 4.5)
 & (5.5, 6.0)
 & (\underline{2.5}, \underline{2.3})   
 & (3.0, 3.5)
 & (2.8, 2.8)
 & (\textbf{1.8}, \textbf{2.0}) \\ \bottomrule[1.5pt]
\end{tabular}
}
\end{table}

\noindent\textbf{Zero-shot Anomaly Detection in the Medical Domain.}
We also conduct experiments in the medical domain to further investigate the generalization ability of these ZSAD methods. The results exhibit a similar trend to those in the industrial domain, where methods with training outperform SAA and WinCLIP by a significant margin. AdaCLIP emerges as the top performer with the highest average rankings, showcasing robust generalization capabilities across different domains. As depicted in Fig.~\ref{fig:qualitative}, AdaCLIP demonstrates precise detection of various anomalies across diverse medical categories, such as identifying skin cancer regions in photographic images and detecting thyroid nodules in ultrasound images. AdaCLIP achieves notably superior performance in locating abnormal lesion/tumor regions compared to other ZSAD methods. \rv{More quantitative and qualitative results in Appendix Section 5-7 further illustrate the superior ZSAD performance of AdaCLIP.}

\subsection{Ablation Study}

\begin{figure}[t]
  \centering
  \includegraphics[width=0.98\linewidth]{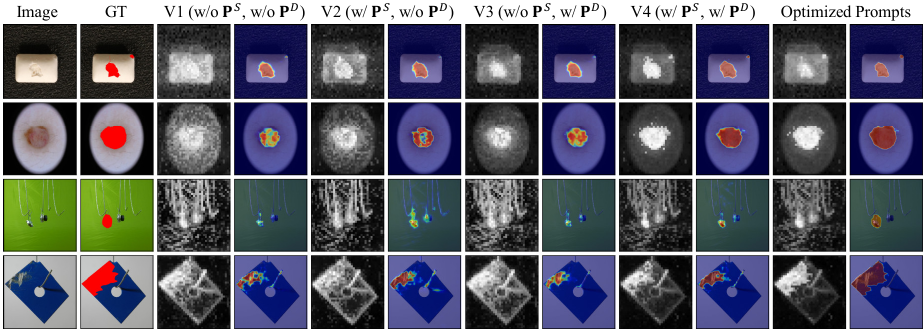}
  \caption{\textbf{Visualization of Patch Embeddings and Anomaly Maps under Different Prompts.} PCA is utilized to reduce the dimension of patch embeddings for enhanced visualization. For individual models, the left shows patch embeddings and the right displays anomaly maps.}
  \label{fig:prompts}
\end{figure}

\begin{figure}[t]
\begin{minipage}[c]{0.60\linewidth}
\captionof{table}{Ablation Results of Static prompts ${\mathbf{P}^S}$ and Dynamic prompts ${\mathbf{P}^D}$.}
\label{tab:prompt}
\resizebox{\textwidth}{!}{
\begin{tabular}{@{}c|cc|cccc@{}}
\toprule[1.5pt]
\multirow{2}{*}{Model} &
\multirow{2}{*}{${\mathbf{P}^S}$}
& 
\multirow{2}{*}{${\mathbf{P}^D}$}  & \multicolumn{2}{c}{Medical Domain} & \multicolumn{2}{c}{Industrial Domain} \\ \cmidrule(l){4-5} \cmidrule(l){6-7}
                       &                    &                    & Image-level    & Pixel-level   & Image-level  & Pixel-level 
\\ \midrule
V1 & \textcolor{gray}{\ding{56}}  & \textcolor{gray}{\ding{56}}  & (87.9, 58.3) & (83.9, 54.3) & (86.7, 85.0) & (92.8, 45.9) \\
V2 & \ding{52} &  \textcolor{gray}{\ding{56}} & (88.8, 60.4) & (84.8, 56.4) & (89.1, 88.7) & (94.1, 48.2) \\
V3 & \textcolor{gray}{\ding{56}}  & \ding{52} & (88.4, 60.0) & (84.4, 57.0) & (86.9, 87.1) & (93.5,  46.1) \\
V4 & \ding{52}  & \ding{52}  & (\textbf{94.6}, \textbf{89.6}) & (\textbf{85.3}, \textbf{57.4}) & (\textbf{90.2}, \textbf{89.6})   & (\textbf{94.2}, \textbf{50.2})   \\ \bottomrule[1.5pt]
\end{tabular}
}

\centering
\captionof{table}{Ablation results on HSF.}
\label{tab:HSF}
\setlength\tabcolsep{6.0pt}
\resizebox{1\textwidth}{!}{
\begin{tabular}{@{}ccccc@{}}
\toprule[1.5pt]
\multirow{2}{*}{HSF}  & \multicolumn{2}{c}{Medical Domain} & \multicolumn{2}{c}{Industrial Domain} \\ \cmidrule(l){2-3} \cmidrule(l){4-5}
                     & Image-level    & Pixel-level   & Image-level  & Pixel-level \\ \midrule
\textcolor{gray}{\ding{56}} & (91.8, 88.7)       & (85.7, 57.7)       & (86.0, 85.8)       & (93.8, 49.9)       \\
\ding{52} & (94.6, 89.6)       & (85.3, 57.4)       & (90.2, 89.6)       & (94.2, 50.2)       \\ \bottomrule[1.5pt]
\end{tabular}
}
\end{minipage}\hfill
\begin{minipage}[c]{0.36\linewidth}
  \input{figures/figure_heatmap}
\end{minipage}
\end{figure}

\begin{figure}[t]
\begin{minipage}[b]{0.46\linewidth}
\captionof{table}{ZSAD Performance in the Medical Domain with Varied Training Data. Top: Image-level AD. Bottom: Pixel-level AD.}
\label{tab:training_data}
\resizebox{\textwidth}{!}{
\begin{tabular}{@{}cccc@{}}
\toprule[1.5pt]
Dataset         & {Medical}   & {Industrial} & {Both}      \\ \midrule
HeadCT   & (76.0, 72.3)         & (81.6, 78.8)               & (\textbf{91.4}, \textbf{85.2}) \\
BrainMRI & (57.6, 76.0)        & (86.4, 85.3)               & (\textbf{94.8}, \textbf{91.2}) \\
Br35H    & (68.7, 68.9)          & (68.8, 69.4)               & (\textbf{97.7}, \textbf{92.4}) \\
Average  & (67.4, 72.4)          & (78.9, 77.8)               & (\textbf{94.6}, \textbf{89.6}) \\ \midrule \midrule
ISIC     & (68.5, 49.5)          & (89.2, \textbf{72.3})      & (\textbf{89.3}, 71.4)          \\
ColonDB  & (89.4, 55.4)          & (78.5, 31.3)               & (\textbf{90.4}, \textbf{58.2}) \\
ClinicDB & (\textbf{91.3}, \textbf{65.1}) & (78.2, 39.5)               & (84.4, 58.2)          \\
TN3K     & (69.7, 33.6)          & (75.9, 41.4)               & (77.2, \textbf{41.9}) \\
Average  & (79.7, 50.9)          & (80.5, 46.1)               & (\textbf{85.3}, \textbf{57.4}) \\ \bottomrule[1.5pt]
\end{tabular}
}
\end{minipage}\hfill
\begin{minipage}[b]{0.50\linewidth}
  \input{figures/figure_training_data}
\end{minipage}
\end{figure}

\noindent\textbf{Influence of Prompts.} Table~\ref{tab:prompt} presents the detection performance of AdaCLIP with different combinations of static prompts and dynamic prompts, namely V1 (w/o ${\mathbf{P}^S}$, w/o ${\mathbf{P}^D}$), V2 (w/ ${\mathbf{P}^S}$, w/o ${\mathbf{P}^D}$), V3 (w/o ${\mathbf{P}^S}$, w/ ${\mathbf{P}^D}$), and V4 (w/ ${\mathbf{P}^S}$, w/ ${\mathbf{P}^D}$). The superior performance of V2 and V3 to V1 shows that both prompts are useful. V4 with hybrid prompts brings the most significant improvements. This is because static prompts struggle to capture diverse anomalies, while solely dynamic prompts are not sufficient. The combined hybrid prompts offer robust and flexible adaptation, thereby offering better ZSAD performance.  Fig.~\ref{fig:prompts} visualizes the patch embeddings and anomaly maps to delve into the influence of prompts. It clearly shows that both prompts are useful in highlighting the abnormal patch embeddings, facilitating precise predictions. However, with solely static or dynamic prompts, the prediction results are not perfect. In comparison, the model (V4) with hybrid prompts can detect anomalies more accurately.   \rv{We also evaluate the influence of multimodal prompts and find it crucial to prompt both the text and image encoders, as shown in Appendix Section 2.1.
}

\noindent\textbf{Analysis on Prompting Depth and Length.} Fig.~\ref{fig:heatmap} visualizes the ZSAD performance of AdaCLIP under different prompting depths ($J$) and prompting lengths ($M$). 
Significantly, the performance of AdaCLIP does not exhibit continuous improvement with larger $J$ and $M$. This is because the incorporation of more learnable prompting parameters introduces a risk of overfitting the auxiliary training dataset. To mitigate this, we employ the default setting $J=4$ and $M=5$, ensuring consistently high AD performance across both domains.

\noindent\textbf{Influence of HSF.} Table~\ref{tab:HSF} showcases the impact of HSF. The results reveal a significant improvement of image-level ZSAD performance across both medical and industrial domains with the introduction of HSF compared to maximum-based image-level AD (without HSF). For instance, the image-level AUROC increases from 86.0\% to 90.2\% in the industrial domain. This improvement is attributed to the ability of HSF to aggregate the semantics of abnormal regions from multiple hierarchies. Conversely, relying on the maximum value of anomaly maps for image-level AD yields suboptimal results. \rv{Additional analysis of HSF is provided in Appendix Section 2.2.
}

\noindent\textbf{Influence of Annotated Auxiliary Data.} We conduct experiments in the medical domain to explore the influence of annotated auxiliary data, as illustrated in Table~\ref{tab:training_data} and Fig.~\ref{fig:training_data}. 
Relying exclusively on medical datasets for training results in subpar ZSAD performance, as illustrated by the notable underperformance on ISIC when trained solely with medical data.
This can be attributed to the lack of data diversity within the selected medical dataset, such as ColonDB~\cite{ColonDB} or ClinicDB~\cite{ClinicDB}. The utilized industrial datasets offer more diverse anomalies, thereby providing greater generalization capacity when trained with them. Notably, training with ColonDB brings surprisingly promising results on ClinicDB, even surpassing more diverse training sets. This is because ColonDB and ClinicDB both focus on colon polyp detection and thus, these two datasets share similarities despite being acquired through different imaging techniques, as shown in Fig.~\ref{fig:training_data}. Generally, using more diverse auxiliary training sets can improve the generalization ability.

\noindent\textbf{Influence of Backbones.} Table~\ref{tab:backbone} illustrates the impact of different backbones. AdaCLIP demonstrates significantly improved results in both medical and industrial domains with a larger backbone, ViT-L/14@336px, compared to ViT-B/16. Moreover, the additional parameters are lightweight compared to the original CLIP parameters, comprising only 4.6\% (40.7 MB) of the original parameters added to ViT-L/14@336px (890.8 MB). This effectively demonstrates that existing VLMs can be adapted to ZSAD using lightweight parameters.

\subsection{Analysis}

\begin{figure}[t]
\begin{minipage}[c]{0.52\linewidth}

\captionof{table}{Comparison between various backbones. Sizes of original CLIP and added parameters by AdaCLIP parameters are reported in Mega Bytes.}
\label{tab:backbone}
\resizebox{\textwidth}{!}{
\begin{tabular}{@{}cccc@{}}
\toprule[1.5pt]
\multicolumn{2}{c}{Backbone}                                                    & ViT-B/16      & ViT-L/14@336px \\ \midrule
Size & \begin{tabular}[c]{@{}c@{}}(Ori., Added)\end{tabular} & (334.6, 19.6) & (890.8, 40.7)  \\ \midrule
\multirow{2}{*}{\begin{tabular}[c]{@{}c@{}}Industrial\\      Domain\end{tabular}} & \begin{tabular}[c]{@{}c@{}}Image-level\end{tabular} & (81.3, 78.3) & (94.6, 89.6) \\
     & \begin{tabular}[c]{@{}c@{}}Pixel-level\end{tabular}                 & (82.5, 52.7)  & (85.3, 57.4)   \\ \midrule
\multirow{2}{*}{\begin{tabular}[c]{@{}c@{}}Medical \\      Domain\end{tabular}}    & \begin{tabular}[c]{@{}c@{}}Image-level\end{tabular} & (83.9, 84.6) & (90.2, 89.6) \\
     & \begin{tabular}[c]{@{}c@{}}Pixel-level\end{tabular}                 & (91.7, 42.1)  & (94.2, 50.2)   \\ \bottomrule[1.5pt]
\end{tabular}
}

\end{minipage}\hfill
\begin{minipage}[c]{0.44\linewidth}
  \input{figures/figure_feature_tsne}
\end{minipage}
\end{figure}

\noindent\textbf{Rationale behind the ZSAD Scheme with Auxiliary Data.} The ZSAD scheme with auxiliary data successfully tailors existing VLMs, including DINOV2, SAM, and CLIP, for ZSAD. To explore the reason why training with auxiliary data can improve ZSAD performance, we visually analyze the distributions of patch embeddings from these models across two datasets featuring diverse anomalies, \ie, MVTec and VisA. In Fig.~\ref{fig:feature_tsne}, it becomes evident that abnormal patch embeddings in both MVTec and VisA exhibit distinctive characteristics compared to the normal ones. 
Meanwhile, the normal embeddings in these two datasets exhibit similar distributions. Consequently, the decision boundary learned in MVTec is applicable to VisA despite not being trained on VisA. This phenomenon can be attributed to the high-level similarities in normalities and abnormalities present in both datasets as perceived by VLMs. The awareness of these similarities can be harnessed by learning annotated auxiliary data. 

\noindent\textbf{Enhancing ZSAD Performance through Prompt Optimization.} The influence of prompting tokens on predictions becomes apparent in both Table~\ref{tab:prompt} and Fig.~\ref{fig:prompts}. While prompts generated by AdaCLIP are promising, the potential for further improving ZSAD performance exists through prompt optimization. We leverage model V4 in Table~\ref{tab:prompt} and refine its prompts for specific images using corresponding anomaly masks for training. The results are depicted in the right two columns of Fig.~\ref{fig:prompts}, illustrating that optimized prompts result in more discernible abnormal patch embeddings and finer anomaly maps, particularly noticeable in the bottom two rows. This underscores the significance of devising methods to generate optimal prompts tailored to individual images.

\section{Conclusion}
\label{sec:conclusion}
In this study, we introduce AdaCLIP, a generic ZSAD model to detect anomalies across arbitrary novel categories without any reference image. AdaCLIP leverages annotated auxiliary AD data for training and effectively adapts pre-trained CLIP for ZSAD by integrating learnable hybrid prompts. Additionally, a HSF module is proposed to extract region-level anomaly information to enhance image-level AD performance. Through extensive experimentation across 14 datasets spanning industrial and medical domains, AdaCLIP demonstrates promising AD performance in novel categories from different domains. 

\noindent\textbf{Discussion and Limitations.} \rv{Our experimental results demonstrate the potential of AdaCLIP as a powerful solution for ZSAD. We believe that leveraging more diverse annotated auxiliary anomaly detection datasets can improve the generalization capability of AdaCLIP. In fact, like any other ZSAD method, AdaCLIP might fail when testing data that significantly depart from auxiliary training data, as shown in Sec. 7.1 in the Appendix. 
}


\section*{Acknowledgements}
This paper is supported in part by FAIR (Future Artificial Intelligence Research) project, funded by the NextGenerationEU program within the PNRR-PE-AI scheme (M4C2, Investment 1.3, Line on Artificial Intelligence), in part by the Ministry of Industry and Information Technology of the People's Republic of China under Grant \#2023ZY01089, and in part by the China Scholarship Council (CSC) under Grant 202306160078. 





%
%
\bibliographystyle{splncs04}



\section*{Supplementary material}

# Supplementary Materials for AdaCLIP: Adapting CLIP with Hybrid Learnable Prompts for Zero-Shot Anomaly Detection

In this appendix, we present more details about the dataset (Section 1), the proposed AdaCLIP (Section 2), and the selected baselines (Section 3). Section 4 provides a fair and comprehensive comparison between the proposed AdaCLIP and another popular ZSAD method, AnomalyCLIP. Section 5 presents comparison results between the proposed AdaCLIP and other popular full-shot unsupervised anomaly detection methods, demonstrating the potential practical applicability of the proposed AdaCLIP. Sections 6 and 7 offer more quantitative and qualitative comparisons.

## 1 Dataset Details

In this study, we conduct extensive experiments on 14 public datasets covering industrial and medical domains across three modalities (photography, radiology, and endoscopy) to assess the effectiveness of our methods. We solely utilize the test data from these datasets, and their relevant information is presented in Table 1. We default to using two datasets, MVTec AD [2] and ClinicDB [3], as auxiliary data for training. Additionally, for evaluations on MVTec AD and ClinicDB, we employ VisA [17] and ColonDB [14] for training.

**Table 1: Key statistics the utilized datasets.** $|\mathcal{C}|$ denotes to the number of categories in individual datasets.

| Domain | Anomaly Detection | | Dataset | Category | Modality | $|\mathcal{C}|$ | Normal and anomalous samples |
| --- | --- | --- | --- | --- | --- | --- | --- |
| | Image-level | Pixel-level | | | | | |
| Industrial | ✔ | ✔ | MVTec AD | Obj & Texture | Photography | 15 | (467,1258) |
| | ✔ | ✔ | VisA | Obj | Photography | 12 | (962,1200) |
| | ✔ | ✔ | MPDD | Obj | Photography | 6 | (176,282) |
| | ✔ | ✔ | BTAD | Obj | Photography | 3 | (451,290) |
| | ✔ | ✔ | KSDD | Obj | Photography | 1 | (181,74) |
| | ✔ | ✔ | DAGM | Texture | Photography | 10 | (6996,1054) |
| | ✔ | ✔ | DTD-Synthetic | Texture | Photography | 12 | (357,947) |
| Medical | ✔ | ✘ | HeadCT | Brain | Radiology (CT) | 1 | (100,100) |
| | ✔ | ✘ | BrainMRI | Brain | Radiology (MRI) | 1 | (98,155) |
| | ✔ | ✘ | Br35H | Brain | Radiology (MRI) | 1 | (1500,1500) |
| | ✘ | ✔ | ISIC | Skin | Photography | 1 | (0,379) |
| | ✘ | ✔ | ClinicDB | Colon | Endoscopy | 1 | (0,612) |
| | ✘ | ✔ | ColonDB | Colon | Endoscopy | 1 | (0,380) |
| | ✘ | ✔ | TN3K | Thyroid | Radiology (Ultrasound) | 1 | (0,614) |

## 2   Module Details

### 2.1   Hybrid Learnable Prompts

We introduce hybrid learnable prompts for adapting the pre-trained CLIP [12] for the ZSAD task. Figure 1 presents the details of hybrid learnable prompts. In particular, we utilize a pre-trained and frozen CLIP image encoder to extract image embeddings that contain high-level semantic information. Then for image and text encoders, we employ a simple linear layer to project the image embeddings into dynamic prompts, respectively. These dynamic prompts are then summed with static prompts from the initial $J$ layers as final hybrid prompts for the image and text encoders. While CoCoOp [15] employs a similar design of hybrid (static+dynamic) prompts, our proposed AdaCLIP prompts both image and text encoders for improved adaptation. We further examine the impact of prompting encoders, with results presented in Table 2. The data indicates that solely prompting the text encoder, as CoCoOp does, results in a performance decrease of 7.2% (0.1%) in image (pixel)-level AUROCs for the medical domain and 2.0% (0.7%) for the industrial domain. Therefore, our multimodal prompting approach more effectively leverages the multimodal capabilities of CLIP, enhancing its potential for zero-shot anomaly detection.

### 2.2   Hybrid Semantic Fusion

Previous maximum value-based image-level anomaly detection methods [4, 8] may exhibit sensitivity to prediction noise. In contrast, we propose a Hybrid Semantic Fusion (HSF) module aimed at fusing region-level anomalies into a semantic-rich image embedding to enhance image-level anomaly detection performance. Specifically, patch embeddings from individual hierarchies are clustered using the KMeans++ algorithm [1]. We hypothesize that these clusters should represent different regions within the image, with clusters having the highest average anomaly scores likely corresponding to abnormal regions. To validate this assumption, we visualize the clustering results in Fig. 2. It is apparent that the clusters delineate distinct regions within the image. Also, the cluster with the highest average anomaly score typically denotes the abnormal region. Consequently, the HSF module aggregates the centroids of these clusters with the highest average anomaly scores into the semantic-rich image embedding, which encapsulates multi-hierarchy context pertaining to region-level anomalies, thereby significantly enhancing image-level anomaly detection. As shown in Table 3, we investigated anomaly detection performance with varying $K \in [10, 20, 40, 80]$. HSF consistently enhances image-level detection results compared to the maximum value-based method, achieving improvements of 2.4% (2.9%), 2.8% (4.2%), 3.8% (2.4%), and 0.3% (4.1%) in image-level AUROCs for medical (industrial) domains, respectively. A larger $K$ results in smaller clusters, and when clusters become sufficiently small, HSF degrades to the maximum value-based method. Optimal

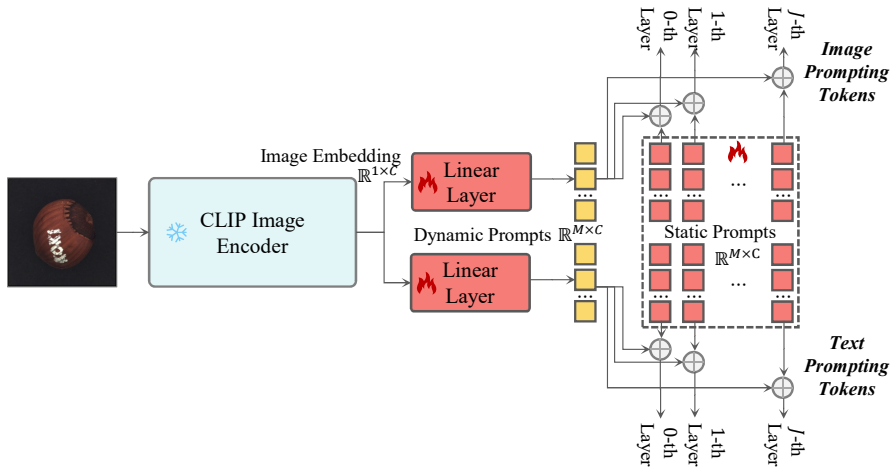

**Fig. 1: Illustration of Hybrid Learnable Prompts.** This illustration depicts the utilization of two linear layers in conjunction with a shared pre-trained CLIP image encoder to generate dynamic prompts for both the image and text encoders. These dynamic prompts, along with the static prompts from the initial $J$ layers, are then combined to prompt the encoders effectively.

**Table 2:** Influence of prompting encoders. The best performance is in **bold**.

| Prompting Encoder | | Medical Domain | | Industrial Domain | |
|---|---|---|---|---|---|
| Image | Text | Image-level | Pixel-level | Image-level | Pixel-level |
| ✔ | ✘ | (86.6, 49.9) | (80.6, 42.9) | (87.6, 85.1) | (93.9, 48.2) |
| ✘ | ✔ | (87.4, 59.0) | (85.2, 57.0) | (88.2, 86.9) | (93.5, 49.8) |
| ✔ | ✔ | (**94.6**, **89.6**) | (**85.3**, **57.4**) | (**90.2**, **89.6**) | (**94.2**, **50.2**) |

performance is ideally obtained when clusters match the size of testing anomalies. However, due to the variability in anomaly sizes across testing categories and samples, achieving an optimal $K$ for both medical and industrial domains is challenging, as indicated in Table 3. Therefore, we set $K = 20$ by default.

## 3 Comparison Method Details

We compare the proposed AdaCLIP with several SOTA methods. Table 4 highlights the key differences between these methods. Notably, AnomalyGPT [7] and AnomalyCLIP [16] are the most relevant concurrent works. In comparison to AdaCLIP, AnomalyGPT uses learnable static prompts but lacks zero-shot anomaly detection capability. While AnomalyCLIP also utilizes prompting learning to enhance ZSAD performance, it solely adds static prompts to the text

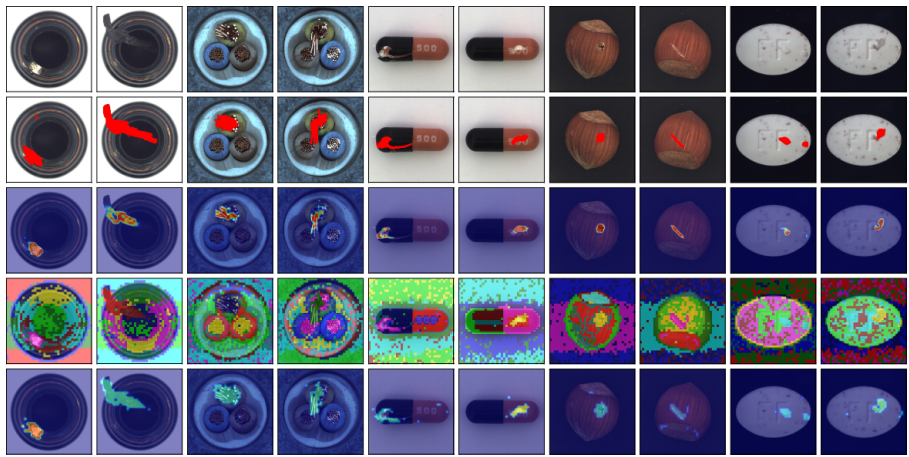

**Fig. 2: Illustration of HSF.** Top to bottom: input images, ground truths, anomaly maps, clustering results, and clusters with average anomaly scores. It clearly demonstrates that HSF can identify abnormal regions and then extract region-level features. The resulting semantic-rich image embedding comprises multi-hierarchy region-level features, enhancing robust image-level anomaly detection.

**Table 3:** Ablation on the number ($K$) of clusters in HSF.

| HSF | Medical Domain | | Industrial Domain | |
|---|---|---|---|---|
| | Image-level | Pixel-level | Image-level | Pixel-level |
| ✗ | (91.8, 88.7) | (85.7, 57.7) | (86.0, 85.8) | (93.8, 49.9) |
| $K$=10 | (94.2, <u>89.6</u>) | (<u>86.9</u>, **59.4**) | (88.9, 87.3) | (93.9, 50.1) |
| $K$=20 | (<u>94.6</u>, <u>89.6</u>) | (85.3, 57.4) | (**90.2**, **89.6**) | (**94.2**, <u>50.2</u>) |
| $K$=40 | (**95.6**, **90.5**) | (85.4, 56.0) | (88.4, 87.8) | (93.3, 48.4) |
| $K$=80 | (92.1, 88.9) | (**87.1**, <u>58.9</u>) | (<u>90.1</u>, <u>89.4</u>) | (<u>94.1</u>, **50.8**) |

encoder of CLIP. In the proposed AdaCLIP, both static and dynamic prompts for both text and image encoders are developed. Due to the unavailability of a publicly accessible implementation for AnomalyCLIP, we report the comparison results against AnomalyCLIP in Section 4. Implementation and reproduction details of other comparison methods are given as follows:

- SAA [5]: SAA is a novel ZSAD model that integrates Grounding DINO [10] and SAM [9] for anomaly detection without any training. Various manual prompts can be adjusted to enhance ZSAD performance. In the case of MVTec AD and VisA datasets, we adhere to the officially provided prompts [1]. For datasets not covered in the original implementation, default prompts in SAA are utilized for ZSAD.
- WinCLIP [8]: WinCLIP represents a SOTA ZSAD method. It devises an extensive array of manual text prompts tailored specifically for anomaly

---
[1] https://github.com/caoyunkang/Segment-Any-Anomaly

6detection and employs a window scaling strategy for anomaly segmentation. We strictly adhere to the text prompts outlined in the original paper.
– APRIL-GAN [6]: APRIL-GAN enhances WinCLIP by employing training on auxiliary AD datam. We adopt the official implementation [2] and adhere to the training settings outlined in the paper, specifically training on both industrial and medical datasets concurrently to improve generalization ability.
– DINOV2 [11]: DINOV2 represents a recent advancement in visual foundation models. We adapt DINOV2 for the ZSAD task by training on auxiliary data like AdaCLIP. Specifically, we utilize the ViT-S/14 architecture[3] as the backbone. Similar to AdaCLIP, we incorporate additional learnable projection layers after the multi-hierarchy patch embeddings and employ the same training set for optimizations. Patch embeddings from the 3rd, 6th, 9th, and 12th layers are selected for multi-hierarchy representations.
– SAM [9]: SAM is recognized as another prominent visual foundation model, primarily crafted for image segmentation tasks. We repurpose SAM for ZSAD by training on auxiliary AD data as well. Specifically, we discard the prompting encoder and mask decoder of SAM and only utilize the backbone of ViT-L architecture[4] for patch embedding extraction. Similar to AdaCLIP, we append trainable projection layers to the patch embeddings from the 6th, 12th, 18th, and 24th layers.
– **AdaCLIP**: As mentioned in the main body, we use the publicly available pre-trained CLIP (ViT-L/14@336px)[5] as the default backbone. We apply the data pre-processing pipeline officially given by CLIP to all images.

Table 4: **Comparison between ZSAD-related methods.** The proposed AdaCLIP introduces both static and dynamic prompts for the text and image encoders for enhanced ZSAD performace.

| Method | Zero-shot Capacity | Supervised Training | Manual Prompts | Learnable Prompts | | Prompting Encoder | |
|---|---|---|---|---|---|---|---|
| | | | | Static | Dynamic | Text | Image |
| AnomalyGPT [7] | ✗ | ✔ | ✔ | ✔ | ✗ | ✔ | ✗ |
| SAA [5] | ✔ | ✗ | ✔ | ✗ | ✗ | ✗ | ✗ |
| WinCLIP [8] | ✔ | ✗ | ✔ | ✗ | ✗ | ✔ | ✗ |
| APRIL-GAN [6] | ✔ | ✔ | ✔ | ✗ | ✗ | ✗ | ✗ |
| AnomalyCLIP [16] | ✔ | ✔ | ✔ | ✔ | ✗ | ✔ | ✗ |
| DINOV2 [11] | ✔ | ✔ | ✗ | ✗ | ✗ | ✗ | ✗ |
| SAM [9] | ✔ | ✔ | ✗ | ✗ | ✗ | ✗ | ✗ |
| **AdaCLIP** | ✔ | ✔ | ✔ | ✔ | ✔ | ✔ | ✔ |

---

[2] https://github.com/ByChelsea/VAND-APRIL-GAN
[3] https://github.com/facebookresearch/dinov2
[4] https://github.com/facebookresearch/segment-anything
[5] https://github.com/mlfoundations/open_clip

**Table 5:** Comparisons between frozen and learnable projection layers. The best performance is in **bold**.

| Proj | Medical Domain | | Industrial Domain | |
|---|---|---|---|---|
| | Image-level | Pixel-level | Image-level | Pixel-level |
| Frozen | (84.4, 81.1) | (79.3, 47.4) | (82.9, 82.6) | (90.1, 41.0) |
| Learnable | (**94.6**, **89.6**) | (**85.3**, **57.4**) | (**90.2**, **89.6**) | (**94.2**, **50.2**) |

## 4  Comparison with AnomalyCLIP

AnomalyCLIP [16] represents a concurrent ZSAD method, introducing learnable object-agnostic prompts for ZSAD, under the assumption of the existence of generic normality and abnormality in an image from whatever category. Due to differences in experimental settings between AnomalyCLIP and our study, as well as the unavailability of publicly available code for AnomalyCLIP (before our submission date), we opt to evaluate AdaCLIP within the framework of AnomalyCLIP for fair comparisons. Specifically, we employ MVTec [2] as the default auxiliary dataset, whereas evaluations on MVTec AD are conducted using VisA [17] for training. The results are depicted in Table 6 and Table 7. The results clearly demonstrate that the proposed AdaCLIP outperforms AnomalyCLIP in average image-level anomaly detection performance across both industrial and medical domains, primarily attributed to the proposed Hybrid Semantic Fusion module. It should be noted that AdaCLIP slightly lags behind AnomalyCLIP in pixel-level detection performance, as AnomalyCLIP incorporates specific design elements to enhance pixel-level anomaly localization, such as a Diagonally Prominent Attention Map mechanism, V-V self-attention, and improved loss functions. In summary, AdaCLIP achieves comparable performance to AnomalyCLIP, while also providing a more thorough investigation into learnable prompts and emphasizing the importance of tailored prompts for individual images.

In addition, we found that AnomalyCLIP differs from AdaCLIP regarding the design of projection layers. Specifically, AnomalyCLIP utilizes the pre-trained and frozen projection layer from CLIP, whereas our proposed AdaCLIP introduces learnable projection layers. To study their differences, we replaced the original learnable projection layers with frozen pre-trained layers, and the comparison results are presented in Table 5. The results clearly show that frozen projection layers lead to significant drops in all metrics. We attribute these drops to the smaller number of learnable parameters with frozen layers, which may limit the adaptation of CLIP to zero-shot anomaly detection.

## 5  Comparison with SOTA Full-shot Methods

In this section, we are interested in the performance gap between AdacLIP and the recently published SOTA full-shot methods, such as PatchCore [13] and

**Table 6:** Comparisons between the proposed AdaCLIP and AnomalyCLIP [16] within the experimental setting of AnomalyCLIP. The results of AnomalyCLIP are directly taken from the original reports. The best performance is in **bold**.

| Metric | Dataset | AnomalyCLIP | AdaCLIP |
| --- | --- | --- | --- |
| Image-level (AUROC) | MVTec AD | **91.5** | 89.6 |
|  | VisA | 82.1 | **83.9** |
|  | MPDD | **77.0** | 76.8 |
|  | BTAD | 88.3 | **88.6** |
|  | KSDD | 84.7 | **94.1** |
|  | DAGM | 97.5 | **98.3** |
|  | DTD | 93.5 | **95.5** |
|  | Average | 87.8 | **89.5** |
| Pixel-level (AUROC) | MVTec AD | **91.1** | 90.3 |
|  | VisA | 95.5 | **95.6** |
|  | MPDD | **96.5** | 96.4 |
|  | BTAD | **94.2** | 92.1 |
|  | SDD | 90.6 | **96.7** |
|  | DAGM | **95.6** | 91.0 |
|  | DTD | **97.9** | 96.9 |
|  | Average | **94.5** | 94.1 |

CDO [4]. Since some datasets do not provide normal training data, we conduct experiments on seven public industrial datasets. As Table 8 shows, AdaCLIP achieves comparable anomaly detection and localization performance compared to PatchCore and CDO, and it even outperforms them in some datasets. This illustrates that AdaCLIP can effectively detect anomalies even in unseen categories by training on auxiliary data. With more extensive data and advanced adapting techniques, future ZSAD methods have opportunities to surpass these SOTO full-shot methods, making ZSAD a viable generic anomaly detection solution.

## 6 Category-Level Quantitative Results

Some datasets contain several categories. In this section, their category-level quantitative results are presented from Table 9 to Table 14 in details.

## 7 Additional Qualitative Results

**Table 7:** Comparisons between the proposed AdaCLIP and AnomalyCLIP [16] within the experimental setting of AnomalyCLIP. The results of AnomalyCLIP are directly taken from the original reports. The best performance is in **bold**.

| Metric | Dataset | AnomalyCLIP | AdaCLIP |
|---|---|---|---|
| Image-level (AUROC) | HeadCT | **93.4** | 91.5 |
| | BrainMRI | 90.3 | **94.8** |
| | Br35H | 94.6 | **97.7** |
| | Average | 92.8 | **94.7** |
| Pixel-level (AUROC) | ISIC | **89.7** | 88.3 |
| | ColonDB | **81.9** | 79.1 |
| | ClinicDB | 82.9 | **84.4** |
| | TN3K | **81.5** | 77.4 |
| | Average | **84.0** | 82.3 |

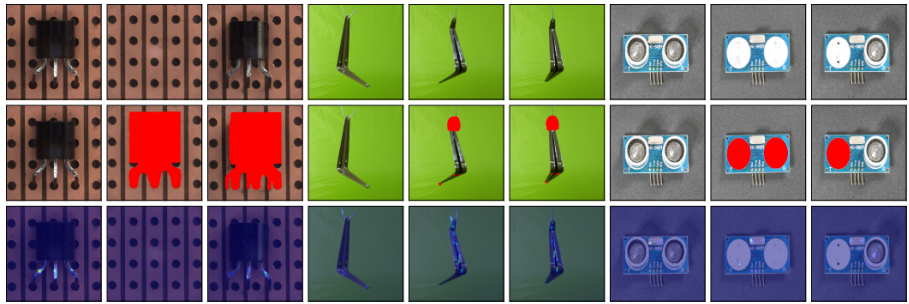

**Fig. 3: Failure cases of AdaCLIP.** Three categories are illustrated with anomaly detection failures. Each category is depicted with its normal state in the left column, and two cases of logical anomalies in the middle and right columns. The top row presents the input images, while the second row shows the ground truth. The bottom row displays the anomaly maps generated by AdaCLIP.

### 7.1 Failure Cases

While the proposed AdaCLIP can achieve promising detection results for arbitrary categories without any references, it may fail to detect anomalies lacking structural deviations. Specifically, the anomalies depicted in Figure 3 exhibit no evident structural deviations. Their abnormality stems from their departure from the expected contextual norms, such as the normal positioning of transistors, among others. However, detecting these anomalies without references poses significant challenges. In the future, it may be worthwhile to explore the integration of more intricate textual prompts describing the normal state to enhance the detection of such anomalies in the absence of references.

Table 8: **Comparisons between the proposed ZSAD method AdaCLIP and full-shot unsupervised AD methods PatchCore and CDO.** The best performance is in **bold**, and the second-best is underlined.

| Metric | Dataset | PatchCore [13] | CDO [4] | **AdaCLIP** |
|---|---|---|---|---|
| Image-level (AUROC, max-F1) | MVTec AD | (**98.8**, **98.3**) | (97.1, 97.0) | (89.2, 90.6) |
| | VisA | (92.7, 89.8) | (**95.0**, **91.4**) | (85.8, 83.1) |
| | MPDD | (94.4, 93.5) | (**95.8**, **94.1**) | (76.0, 82.5) |
| | BTAD | (94.4, **96.6**) | (**97.6**, 94.3) | (88.6, 88.2) |
| | SDD | (93.6, 76.4) | (96.0, 83.5) | (**97.1**, **90.7**) |
| | DAGM | (95.0, 93.6) | (95.1, 92.5) | (**99.1**, **97.5**) |
| | DTD | (**97.5**, **96.4**) | (96.8, 95.7) | (95.5, 94.7) |
| | Average | (95.2, 92.1) | (**96.2**, **92.6**) | (90.2, 89.6) |
| Pixel-level (AUROC, max-F1) | MVTec AD | (**98.4**, **62.2**) | (98.2, 60.1) | (88.7, 43.4) |
| | VisA | (98.6, **43.9**) | (**99.0**, 43.5) | (95.5, 37.7) |
| | MPDD | 98.8, **47.7** | (**99.0**, 46.9) | (96.1, 34.9) |
| | BTAD | (97.5, 54.4) | (**98.1**, **60.4**) | (92.1, 51.7) |
| | SDD | (95.6, 36.9) | (**97.9**, 35.7) | (97.7, **54.5**) |
| | DAGM | (97.2, **59.4**) | (**97.3**, 58.3) | (91.5, 57.5) |
| | DTD | (98.2, 56.8) | (**98.3**, 59.9) | (97.9, **71.6**) |
| | Average | (97.7, 51.6) | (**98.2**, **52.1**) | (94.2, 50.2) |

### 7.2 Results in the Industrial Domain

In this section, we provide additional qualitative results in the industrial domain. Further details can be observed in Figure 4 to Figure 23.

### 7.3 Results in the Medical Domain

This section showcases additional qualitative results in the medical domain, spanning from Figure 24 to Figure 26.

AdaCLIP    11

Table 9: Comparisons of ZSAD methods on MVTec AD. The best performance is in **bold**, and the second-best is underlined.

| Metric | Category | w/o supervised training | | w/i supervised training | | | |
|---|---|---|---|---|---|---|---|
| | | SAA [5] | WinCLIP [8] | DINOV2 [11] | SAM [9] | APRIL-GAN [6] | **AdaCLIP** |
| Image-level (AUROC, max-F1) | bottle | (75.5, 89.4) | (99.2, 97.6) | (**99.4**, **98.4**) | (94.6, 96.0) | (92.9, 94.0) | (94.4, 91.6) |
| | cable | (63.7, 76.0) | (86.5, 84.5) | (49.9, 76.0) | (59.6, 76.0) | (62.6, 77.5) | (**90.6**, **87.8**) |
| | capsule | (42.0, 90.5) | (72.9, 91.4) | (65.2, 90.5) | (54.5, 90.5) | (79.1, 90.8) | (**91.5**, **92.4**) |
| | carpet | (99.5, 98.3) | (**100.0**, **99.4**) | (87.6, 90.8) | (80.5, 88.0) | (99.2, 98.3) | (82.1, 86.5) |
| | grid | (83.7, 86.4) | (**98.8**, **98.2**) | (97.7, 96.4) | (90.4, 90.3) | (89.3, 89.3) | (90.0, 90.8) |
| | hazelnut | (83.2, 83.3) | (**93.9**, **89.7**) | (43.8, 77.8) | (58.0, 79.1) | (76.8, 81.2) | (80.2, 82.6) |
| | leather | (99.3, 97.8) | (**100.0**, **100.0**) | (100.0, 99.4) | (90.4, 90.9) | (99.7, 98.9) | (99.8, 99.5) |
| | metal_nut | (34.8, 89.4) | (**97.1**, **96.3**) | (44.3, 89.4) | (60.6, 89.4) | (45.6, 89.4) | (83.5, 90.5) |
| | pill | (50.6, 91.6) | (79.1, 91.6) | (69.7, 91.6) | (68.2, 92.5) | (**90.4**, 92.5) | (82.9, **92.8**) |
| | screw | (46.4, 85.9) | (83.3, 87.4) | (77.5, 85.6) | (68.6, 86.2) | (70.1, 86.3) | (**87.0**, **89.7**) |
| | tile | (95.7, 93.9) | (**100.0**, **99.4**) | (74.2, 83.6) | (41.9, 83.6) | (93.4, 93.9) | (90.5, 91.4) |
| | toothbrush | (22.2, 83.3) | (87.5, 87.9) | (64.0, 83.3) | (68.3, 85.7) | (72.2, 84.9) | (**93.6**, **95.2**) |
| | transistor | (37.0, 57.1) | (**88.0**, **79.5**) | (51.7, 57.1) | (53.3, 57.1) | (72.8, 68.2) | (82.1, 77.5) |
| | wood | (**99.8**, **99.2**) | (99.4, 98.3) | (97.0, 96.6) | (96.4, 96.7) | (96.8, 95.2) | (98.3, 96.7) |
| | zipper | (19.4, 88.2) | (91.5, 92.9) | (**93.9**, **93.9**) | (76.0, 88.5) | (93.4, 92.7) | (91.5, 93.9) |
| | Average | (63.5, 87.4) | (**91.8**, **92.9**) | (74.4, 87.4) | (70.7, 86.0) | (82.3, 88.9) | (89.2, 90.6) |
| Pixel-level (AUROC, max-F1) | bottle | (66.5, 37.7) | (89.5, 58.1) | (81.6, 57.8) | (**90.5**, 51.0) | (80.8, **60.5**) | (90.4, 54.3) |
| | cable | (69.2, **30.0**) | (77.0, 19.7) | (60.7, 9.2) | (76.3, 18.1) | (71.7, 18.8) | (**79.8**, 19.6) |
| | capsule | (62.1, 17.4) | (86.9, 21.7) | (80.3, 23.4) | (**90.4**, 16.0) | (72.8, 29.6) | (82.3, **31.8**) |
| | carpet | (83.7, 57.8) | (95.4, 49.7) | (**99.3**, **72.6**) | (92.9, 43.5) | (97.1, 70.8) | (97.3, 61.4) |
| | grid | (63.3, 25.5) | (82.2, 18.6) | (92.3, 35.5) | (86.5, 25.1) | (84.6, 34.1) | (**96.9**, **43.7**) |
| | hazelnut | (89.8, 47.1) | (94.3, 37.6) | (91.7, 23.0) | (92.8, 26.9) | (95.8, 32.8) | (**97.8**, **51.8**) |
| | leather | (89.7, **68.8**) | (96.7, 39.7) | (98.3, 54.3) | (89.3, 40.1) | (99.0, 56.5) | (**99.2**, 53.4) |
| | metal_nut | (64.0, **36.1**) | (61.0, 32.4) | (67.2, 32.6) | (**75.5**, 33.4) | (65.5, 28.8) | (74.3, 34.8) |
| | pill | (**91.7**, **53.6**) | (80.0, 17.6) | (90.4, 32.8) | (88.3, 24.0) | (87.0, 36.6) | (86.4, 37.6) |
| | screw | (68.8, 15.0) | (89.6, 13.5) | (98.1, **52.7**) | (97.0, 25.5) | (96.4, 22.1) | (**98.4**, 41.9) |
| | tile | (86.6, **71.0**) | (77.6, 32.6) | (82.6, 56.8) | (61.2, 21.9) | (80.2, 63.3) | (**88.5**, 61.9) |
| | toothbrush | (66.8, 8.0) | (86.9, 17.1) | (87.6, 11.5) | (85.7, 10.4) | (90.6, 21.6) | (**94.9**, **31.9**) |
| | transistor | (66.9, 20.1) | (**74.7**, **30.5**) | (65.2, 17.1) | (70.6, 17.2) | (60.5, 16.5) | (63.2, 17.5) |
| | wood | (84.3, 63.0) | (93.4, 51.5) | (**95.9**, 62.8) | (91.7, 57.1) | (89.1, **64.2**) | (87.9, 57.9) |
| | zipper | (78.4, 19.7) | (91.6, 34.4) | (**97.1**, 51.3) | (92.3, 30.5) | (84.3, 41.2) | (93.8, **52.1**) |
| | Average | (75.5, 38.1) | (85.1, 31.6) | (85.9, 39.6) | (85.4, 29.4) | (83.7, 39.8) | (**88.7**, **43.4**) |

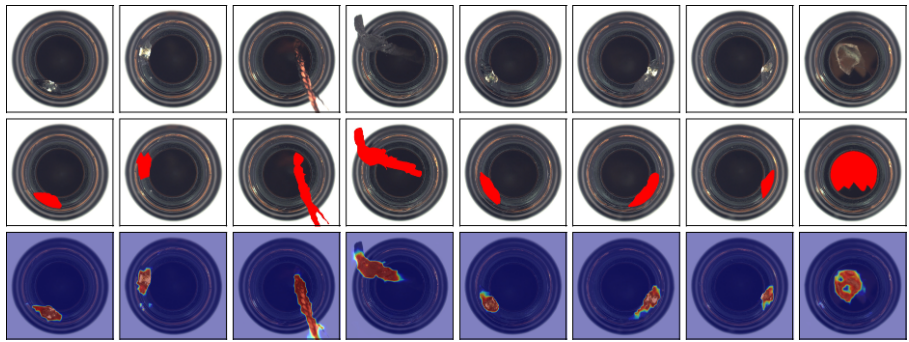

**Fig. 4: Visualization of anomaly maps generated by AdaCLIP for the bottle category in MVTec AD.** The first row displays the input images, while the second row shows the ground truth. The bottom row illustrates the anomaly maps generated by AdaCLIP.

Table 10: Comparisons of ZSAD methods on VisA. The best performance is in **bold**, and the second-best is underlined.

| Metric | Category | w/o supervised training | | w/i supervised training | | | |
|---|---|---|---|---|---|---|---|
| | | SAA [5] | WinCLIP [8] | DINOV2 [11] | SAM [9] | APRIL-GAN [6] | **AdaCLIP** |
| Image-level (AUROC, max-F1) | candle | (63.8, 68.6) | (95.4, 89.4) | (81.3, 75.4) | (58.4, 68.1) | (81.0, 74.7) | (**96.0**, **90.2**) |
| | capsules | (58.1, 76.9) | (85.0, 83.9) | (**92.1**, 88.4) | (38.3, 77.5) | (91.8, **89.2**) | (85.1, 82.1) |
| | cashew | (87.0, 86.1) | (**92.1**, **88.4**) | (56.0, 80.0) | (43.0, 80.0) | (89.4, 88.3) | (91.8, 86.6) |
| | chewinggum | (91.9, 88.4) | (96.5, 94.8) | (85.4, 85.8) | (78.7, 80.0) | (**97.0**, **95.9**) | (96.4, 94.8) |
| | fryum | (39.6, 80.0) | (80.3, 82.7) | (75.5, 84.1) | (71.7, 81.3) | (78.1, 83.2) | (**93.0**, **91.0**) |
| | macaroni1 | (88.7, 82.2) | (76.2, 74.2) | (89.1, **83.4**) | (50.1, 66.7) | (82.2, 78.0) | (**91.6**, 83.1) |
| | macaroni2 | (67.3, 67.6) | (63.7, 69.8) | (**78.7**, **72.6**) | (47.7, 66.7) | (58.7, 66.7) | (64.1, 70.3) |
| | pcb1 | (53.4, 66.9) | (73.6, 71.0) | (65.0, 73.1) | (69.3, 71.2) | (67.5, 68.7) | (**81.1**, **76.9**) |
| | pcb2 | (59.2, 66.7) | (51.2, 67.1) | (54.3, 67.3) | (56.6, 66.7) | (73.9, 71.6) | (**75.3**, **73.7**) |
| | pcb3 | (54.0, 66.5) | (**73.4**, **71.0**) | (57.3, 66.7) | (63.0, 66.9) | (69.1, 67.1) | (64.7, 67.2) |
| | pcb4 | (46.9, 66.5) | (79.6, 74.9) | (72.1, 71.6) | (77.3, 74.1) | (**94.9**, **90.6**) | (93.4, 87.2) |
| | pipe_fryum | (95.8, **94.6**) | (69.7, 80.7) | (96.1, 93.8) | (88.3, 87.3) | (**96.6**, 94.2) | (**96.6**, 94.4) |
| | Average | (67.1, 75.9) | (78.1, 79.0) | (75.2, 78.5) | (61.9, 73.9) | (81.7, 80.7) | (**85.8**, **83.1**) |
| Pixel-level (AUROC, max-F1) | candle | (54.1, 12.8) | (88.9, 22.5) | (98.5, 42.2) | (97.1, 14.6) | (98.5, 41.3) | (**98.9**, **46.6**) |
| | capsules | (81.5, 39.8) | (81.6, 9.2) | (**98.6**, **62.2**) | (88.7, 6.4) | (97.5, 49.0) | (98.6, 52.8) |
| | cashew | (56.4, 13.8) | (84.7, 13.2) | (90.7, 10.9) | (90.2, 13.1) | (92.2, 22.7) | (**95.9**, **39.2**) |
| | chewinggum | (94.9, **83.3**) | (93.3, 41.1) | (99.6, 77.6) | (98.4, 59.3) | (99.4, 78.4) | (99.6, 77.9) |
| | fryum | (92.6, **42.8**) | (88.5, 22.1) | (92.8, 25.7) | (93.4, 26.1) | (93.4, 29.6) | (**94.4**, 30.5) |
| | macaroni1 | (84.1, **42.3**) | (70.9, 7.0) | (98.9, 27.1) | (96.1, 7.4) | (98.8, 29.1) | (**99.5**, 35.0) |
| | macaroni2 | (81.5, **29.9**) | (59.3, 1.0) | (98.0, 21.7) | (95.5, 3.9) | (97.2, 4.6) | (**98.8**, 10.2) |
| | pcb1 | (73.7, **42.1**) | (61.2, 2.4) | (91.3, 10.8) | (89.1, 5.2) | (92.1, 13.1) | (**93.7**, 19.8) |
| | pcb2 | (80.7, 3.5) | (71.6, 4.7) | (**91.3**, 12.1) | (89.3, 8.4) | (90.6, 24.2) | (84.3, **27.7**) |
| | pcb3 | (71.9, 11.2) | (85.3, 10.3) | (89.8, 12.0) | (83.4, 9.4) | (91.0, 23.5) | (**91.8**, **32.2**) |
| | pcb4 | (66.7, 10.2) | (94.4, 32.0) | (94.8, 30.2) | (92.8, 26.9) | (94.7, 37.3) | (**96.1**, **43.3**) |
| | pipe_fryum | (79.7, **47.1**) | (75.4, 12.3) | (96.1, 31.8) | (**97.1**, 38.1) | (96.7, 35.2) | (94.6, 37.4) |
| | Average | (76.5, 31.6) | (79.6, 14.8) | (95.0, 30.3) | (92.6, 18.2) | (95.2, 32.3) | (**95.5**, **37.7**) |

Table 11: Comparisons of ZSAD methods on MPDD. The best performance is in **bold**, and the second-best is underlined.

| Metric | Category | w/o supervised training | | w/i supervised training | | | |
|---|---|---|---|---|---|---|---|
| | | SAA [5] | WinCLIP [8] | DINOV2 [11] | SAM [9] | APRIL-GAN [6] | **AdaCLIP** |
| Image-level (AUROC, max-F1) | bracket_black | (37.2, 74.6) | (40.7, 74.6) | (**70.2**, 79.3) | (59.2, 75.9) | (50.7, 75.2) | (62.0, **81.4**) |
| | bracket_brown | (63.4, 81.0) | (33.2, 79.7) | (42.1, 79.7) | (48.8, 80.0) | (70.9, 80.7) | (**71.3**, **81.7**) |
| | bracket_white | (73.1, 74.1) | (41.8, 67.4) | (57.4, 68.2) | (56.7, 71.6) | (68.0, 71.8) | (**74.7**, **75.0**) |
| | connector | (31.9, 48.3) | (**78.6**, 65.1) | (35.7, 50.0) | (76.0, 61.1) | (48.1, 50.0) | (69.9, **66.3**) |
| | metal_plate | (36.9, 84.5) | (**95.5**, 95.1) | (84.9, 87.9) | (93.3, 91.4) | (65.1, 85.4) | (84.6, **95.4**) |
| | tubes | (13.5, 81.2) | (78.4, 83.1) | (84.3, 84.0) | (44.2, 81.7) | (**93.4**, 93.2) | (93.3, **95.2**) |
| | Average | (42.7, 73.9) | (61.4, 77.5) | (62.4, 74.9) | (63.0, 77.0) | (66.0, 76.0) | (**76.0**, **82.5**) |
| Pixel-level (AUROC, max-F1) | bracket_black | (93.9, 1.8) | (46.4, 0.2) | (**96.9**, **22.3**) | (96.0, 4.4) | (96.5, 13.8) | (93.2, 9.1) |
| | bracket_brown | (66.9, 5.3) | (56.4, 1.4) | (92.0, 13.3) | (89.7, 11.2) | (89.9, 9.0) | (**93.8**, **15.9**) |
| | bracket_white | (97.1, **30.5**) | (72.2, 1.0) | (97.6, 1.9) | (**99.3**, 5.4) | (99.3, 9.0) | (97.1, 3.9) |
| | connector | (71.5, 8.2) | (78.8, 10.7) | (93.2, 14.9) | (93.1, 17.7) | (93.5, 26.2) | (**97.4**, **37.7**) |
| | metal_plate | (73.8, 56.9) | (95.7, 69.7) | (95.9, 72.9) | (**96.9**, **77.2**) | (92.5, 60.8) | (95.8, 72.9) |
| | tubes | (87.3, 10.6) | (77.6, 9.5) | (98.1, 61.5) | (94.0, 16.8) | (**99.0**, 64.8) | (**99.2**, **70.1**) |
| | Average | (81.7, 18.9) | (71.2, 15.4) | (95.6, 31.1) | (94.8, 22.1) | (95.1, 30.6) | (**96.1**, **34.9**) |

Table 12: Comparisons of ZSAD methods on BTAD. The best performance is in **bold**, and the second-best is underlined.

| Metric | Category | w/o supervised training | | w/i supervised training | | | |
|---|---|---|---|---|---|---|---|
| | | SAA [5] | WinCLIP [8] | DINOV2 [11] | SAM [9] | APRIL-GAN [6] | **AdaCLIP** |
| Image-level (AUROC, max-F1) | Class01 | (6.6, 82.4) | (89.3, 87.6) | (80.3, 84.3) | (**96.2**, **93.9**) | (87.3, 88.2) | (91.6, 90.8) |
| | Class02 | (72.3, 93.0) | (72.2, 93.0) | (**88.2**, 94.3) | (76.1, 93.0) | (75.2, 93.0) | (78.0, **94.6**) |
| | Class03 | (**98.2**, **93.7**) | (43.0, 22.2) | (69.5, 29.2) | (96.1, 70.1) | (93.0, 64.7) | (96.3, 79.1) |
| | Average | (59.0, **89.7**) | (68.2, 67.6) | (79.3, 69.3) | (**89.4**, 85.7) | (85.2, 82.0) | (88.6, 88.2) |
| Pixel-level (AUROC, max-F1) | Class01 | (49.6, 6.6) | (84.0, 21.8) | (86.0, 44.0) | (**90.6**, 43.7) | (83.9, 41.2) | (87.1, **55.3**) |
| | Class02 | (73.7, 26.4) | (86.4, 33.1) | (**96.0**, **68.7**) | (94.7, 59.5) | (92.2, 58.3) | (92.9, 59.8) |
| | Class03 | (74.0, 11.5) | (47.5, 0.7) | (93.6, 17.4) | (96.2, 37.4) | (92.3, 15.7) | (**96.2**, **40.1**) |
| | Average | (65.8, 14.8) | (72.6, 18.5) | (91.9, 43.4) | (**93.8**, 46.9) | (89.5, 38.4) | (92.1, **51.7**) |

Table 13: Comparisons of ZSAD methods on DAGM. The best performance is in **bold**, and the second-best is underlined.

| Metric | Category | w/o supervised training | | w/i supervised training | | | |
|---|---|---|---|---|---|---|---|
| | | SAA [5] | WinCLIP [8] | DINOV2 [11] | SAM [9] | APRIL-GAN [6] | **AdaCLIP** |
| Image-level (AUROC, max-F1) | Class1 | (96.2, 90.9) | (68.4, 67.8) | (81.0, 76.5) | (**96.3**, 93.3) | (91.3, 85.6) | (96.2, **93.8**) |
| | Class2 | (**100.0**, **100.0**) | (99.8, 99.0) | (99.4, 98.0) | (**100.0**, **100.0**) | (99.8, 99.0) | (**100.0**, 99.7) |
| | Class3 | (100.0, 99.3) | (99.0, 95.6) | (**100.0**, **100.0**) | (94.2, 89.1) | (**100.0**, **100.0**) | (**100.0**, **100.0**) |
| | Class4 | (36.8, 66.7) | (89.0, 80.7) | (55.9, 67.1) | (45.1, 66.7) | (67.2, 69.2) | (**96.6**, **89.5**) |
| | Class5 | (**100.0**, 99.7) | (95.2, 89.0) | (99.7, 98.0) | (88.0, 83.5) | (99.7, 99.0) | (**100.0**, **100.0**) |
| | Class6 | (72.0, 66.9) | (99.8, 98.7) | (91.0, 85.7) | (86.3, 80.8) | (99.9, 99.0) | (**100.0**, **100.0**) |
| | Class7 | (99.7, 98.2) | (96.3, 90.3) | (99.3, 98.7) | (98.5, 93.8) | (**100.0**, 99.3) | (**100.0**, **100.0**) |
| | Class8 | (**100.0**, **99.7**) | (74.2, 9.9) | (81.7, 74.3) | (73.2, 70.8) | (97.2, 93.7) | (99.3, 97.8) |
| | Class9 | (**100.0**, **100.0**) | (96.4, 90.4) | (99.5, 96.7) | (49.2, 67.2) | (97.8, 94.6) | (99.6, 96.9) |
| | Class10 | (66.6, 66.7) | (98.9, 94.5) | (99.4, 96.9) | (96.8, 90.4) | (81.9, 78.3) | (**99.5**, **97.4**) |
| | Average | (87.1, 88.8) | (91.7, 87.6) | (90.7, 89.2) | (82.7, 83.6) | (93.5, 91.8) | (**99.1**, **97.5**) |
| Pixel-level (AUROC, max-F1) | Class1 | (63.9, 39.4) | (76.0, 12.3) | (84.3, 32.1) | (**90.5**, 42.9) | (83.3, 42.0) | (85.4, **47.6**) |
| | Class2 | (74.9, 55.5) | (80.9, 9.3) | (94.3, 57.2) | (**98.9**, 65.8) | (96.7, 63.9) | (97.8, **66.7**) |
| | Class3 | (57.4, 25.9) | (86.8, 19.4) | (87.7, 59.4) | (87.8, 40.0) | (88.1, 65.5) | (**89.8**, **65.7**) |
| | Class4 | (50.0, 2.7) | (**85.6**, 17.4) | (83.6, 18.5) | (79.4, 13.5) | (79.6, 21.0) | (84.8, **23.9**) |
| | Class5 | (61.3, 37.1) | (83.4, 15.4) | (92.3, 64.4) | (90.2, 42.5) | (92.3, 69.5) | (**95.0**, **69.7**) |
| | Class6 | (73.0, 35.3) | (76.9, 19.6) | (**97.8**, 71.2) | (95.0, 70.6) | (96.8, **79.6**) | (94.5, 77.3) |
| | Class7 | (65.5, 45.7) | (85.9, 24.5) | (92.2, 66.1) | (83.2, 51.0) | (89.9, 70.8) | (**94.1**, **72.1**) |
| | Class8 | (48.7, 16.8) | (69.7, 3.5) | (84.5, 21.9) | (83.8, 17.7) | (**88.1**, 56.4) | (87.2, **60.2**) |
| | Class9 | (61.8, 38.2) | (80.1, 2.0) | (**97.6**, **62.4**) | (80.3, 3.7) | (94.7, **62.4**) | (91.3, 48.3) |
| | Class10 | (70.4, 29.2) | (87.9, 15.1) | (94.5, **66.7**) | (**97.2**, 59.0) | (93.9, 48.1) | (94.7, 43.8) |
| | Average | (62.7, 32.6) | (81.3, 13.9) | (90.9, 52.0) | (88.6, 40.7) | (90.3, **57.9**) | (**91.5**, 57.5) |

**Table 14: Comparisons of ZSAD methods on DTD-Synthetic.** The best performance is in **bold**, and the second-best is underlined.

| Metric | Category | w/o supervised training | | w/i supervised training | | | |
|---|---|---|---|---|---|---|---|
| | | SAA [5] | WinCLIP [8] | DINOV2 [11] | SAM [9] | APRIL-GAN [6] | **AdaCLIP** |
| Image-level (AUROC, max-F1) | Blotchy_099 | (**100.0**, **100.0**) | (99.3, 99.4) | (80.6, 88.9) | (58.3, 88.9) | (**100.0**, **100.0**) | (**100.0**, 99.4) |
| | Fibrous_183 | (99.1, 98.7) | (97.0, 94.9) | (52.5, 88.9) | (64.6, 88.9) | (99.3, 97.6) | (99.9, 99.4) |
| | Marbled_078 | (98.3, 97.5) | (98.4, 97.5) | (66.4, 90.9) | (89.4, 91.7) | (**100.0**, **100.0**) | (99.8, 99.4) |
| | Matted_069 | (99.3, 98.1) | (97.5, 96.1) | (59.3, 88.8) | (46.8, 88.8) | (**99.4**, **98.1**) | (90.9, 93.8) |
| | Mesh_114 | (82.5, 81.6) | (76.0, 82.5) | (**95.0**, **92.1**) | (81.8, 83.1) | (93.0, 91.1) | (83.2, 84.3) |
| | Perforated_037 | (98.4, 98.1) | (99.5, 98.8) | (**100.0**, **100.0**) | (96.3, 96.8) | (97.1, 95.6) | (92.5, 83.6) |
| | Stratified_154 | (96.3, 96.3) | (97.6, 96.2) | (99.0, 98.1) | (98.5, 98.1) | (**100.0**, **100.0**) | (**100.0**, **100.0**) |
| | Woven_001 | (98.1, 96.4) | (95.7, 93.6) | (99.8, 99.3) | (95.2, 91.9) | (**100.0**, **100.0**) | (**100.0**, **100.0**) |
| | Woven_068 | (94.6, 91.6) | (96.6, 94.3) | (96.7, 94.9) | (**99.7**, **99.4**) | (99.2, 97.4) | (91.8, 93.2) |
| | Woven_104 | (90.2, 93.0) | (98.1, 98.1) | (91.1, 94.1) | (**99.9**, **99.4**) | (99.4, 98.1) | (92.6, 91.1) |
| | Woven_125 | (98.9, 97.5) | (99.4, 98.7) | (94.3, 95.0) | (99.9, 99.4) | (99.9), 99.4 | (**100.0**, **100.0**) |
| | Woven_127 | (77.5, 72.9) | (86.1, 78.5) | (94.3, 90.5) | (90.6, 84.0) | (**95.8**, **92.8**) | |
| | Average | (94.4, 93.5) | (95.1, 94.1) | (85.8, 93.5) | (81.9, 91.1) | (**98.1**, **96.8**) | (95.5, 94.7) |
| Pixel-level (AUROC, max-F1) | Blotchy_099 | (84.0, 80.3) | (67.3, 11.4) | (97.0, 60.8) | (97.1, 44.9) | (**99.7**, 77.8) | (99.3, **81.1**) |
| | Fibrous_183 | (81.8, 76.1) | (87.2, 28.2) | (96.0, 45.3) | (94.5, 54.2) | (99.5, **78.8**) | (**99.6**, 67.3) |
| | Marbled_078 | (79.7, 71.7) | (78.0, 14.9) | (95.9, 47.2) | (98.1, 71.0) | (99.6, **78.7**) | (**99.7**, 78.4) |
| | Matted_069 | (70.0, 55.9) | (90.2, 17.8) | (89.5, 22.9) | (84.7, 12.4) | (**99.2**, **72.3**) | (96.9, 67.9) |
| | Mesh_114 | (68.7, 50.8) | (76.1, 9.5) | (96.7, **72.6**) | (93.6, 49.7) | (94.7, 66.1) | (**97.3**, 68.7) |
| | Perforated_037 | (80.9, 59.0) | (76.9, 8.4) | (**99.0**, **75.3**) | (95.1, 67.6) | (95.8, 68.0) | (95.8, 70.0) |
| | Stratified_154 | (81.6, 70.4) | (71.8, 26.9) | (99.1, **81.1**) | (98.0, 71.3) | (99.0, 77.4) | (**99.3**, 66.9) |
| | Woven_001 | (80.0, 70.4) | (83.0, 10.2) | (**99.7**, 77.2) | (98.3, 71.9) | (99.6, 77.7) | (99.5, **78.8**) |
| | Woven_068 | (73.4, 49.8) | (92.1, 21.9) | (98.4, 66.5) | (**99.0**, **76.4**) | (97.5, 71.2) | (96.4, 64.6) |
| | Woven_104 | (84.2, 52.8) | (79.4, 18.2) | (98.4, 66.8) | (98.6, **72.0**) | (96.3, 69.2) | (**98.7**, 70.8) |
| | Woven_125 | (75.3, 64.6) | (84.8, 20.2) | (**99.7**, 82.5) | (99.7, 82.3) | (99.7, 82.3) | (99.5, **83.1**) |
| | Woven_127 | (60.3, 25.5) | (66.7, 6.2) | (**94.6**, **62.1**) | (83.9, 10.0) | (93.1, 53.4) | (93.3, 62.0) |
| | Average | (76.7, 60.6) | (79.5, 16.1) | (97.0, 63.4) | (95.0, 56.7) | (97.8, **72.7**) | (**97.9**, 71.6) |

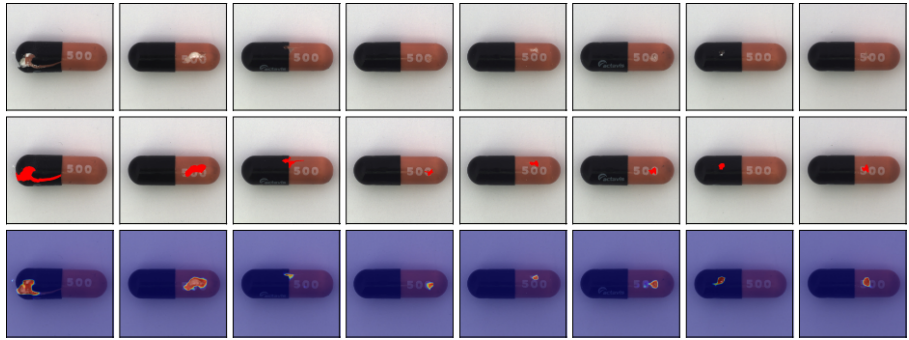

**Fig. 5: Visualization of anomaly maps generated by AdaCLIP for the capsule category in MVTec AD.** The first row displays the input images, while the second row shows the ground truth. The bottom row illustrates the anomaly maps generated by AdaCLIP.

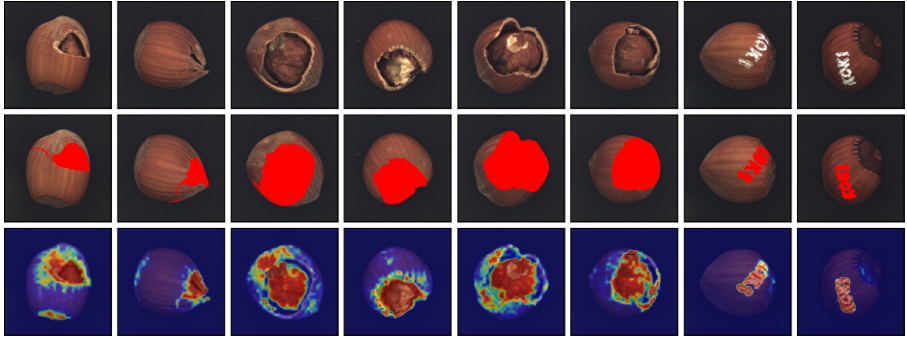

**Fig. 6: Visualization of anomaly maps generated by AdaCLIP for the hazelnut category in MVTec AD.** The first row displays the input images, while the second row shows the ground truth. The bottom row illustrates the anomaly maps generated by AdaCLIP.

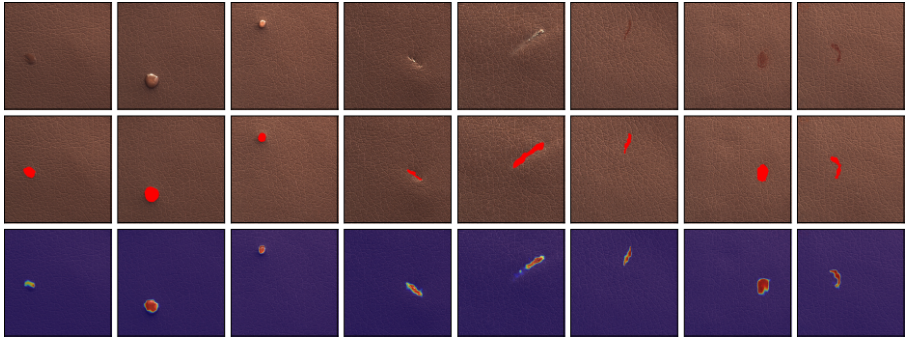

**Fig. 7: Visualization of anomaly maps generated by AdaCLIP for the leather category in MVTec AD.** The first row displays the input images, while the second row shows the ground truth. The bottom row illustrates the anomaly maps generated by AdaCLIP.

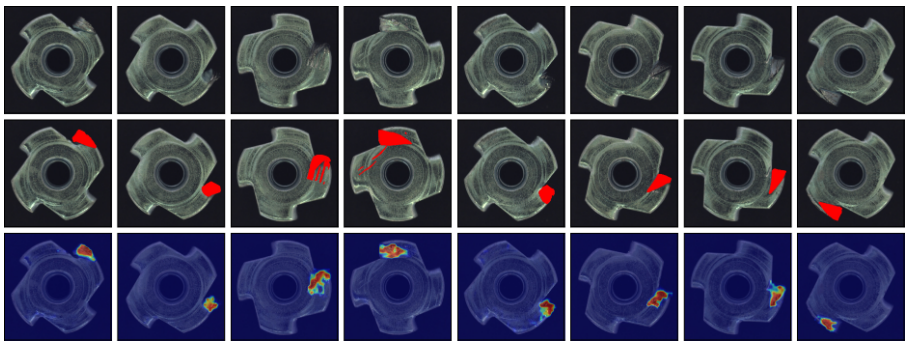

**Fig. 8: Visualization of anomaly maps generated by AdaCLIP for the metal_nut category in MVTec AD.** The first row displays the input images, while the second row shows the ground truth. The bottom row illustrates the anomaly maps generated by AdaCLIP.

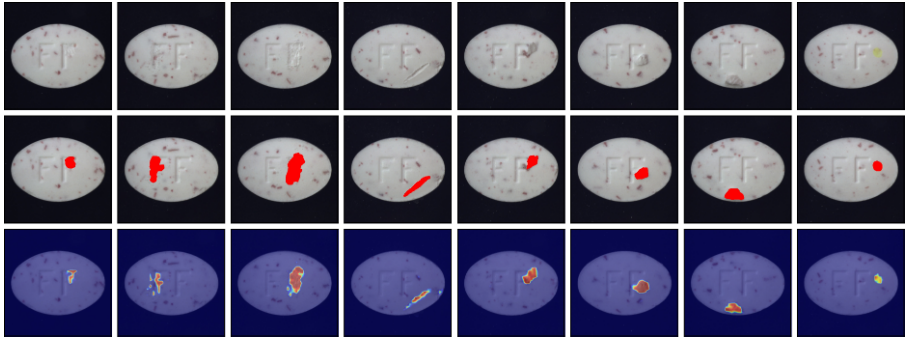

**Fig. 9: Visualization of anomaly maps generated by AdaCLIP for the pill category in MVTec AD.** The first row displays the input images, while the second row shows the ground truth. The bottom row illustrates the anomaly maps generated by AdaCLIP.

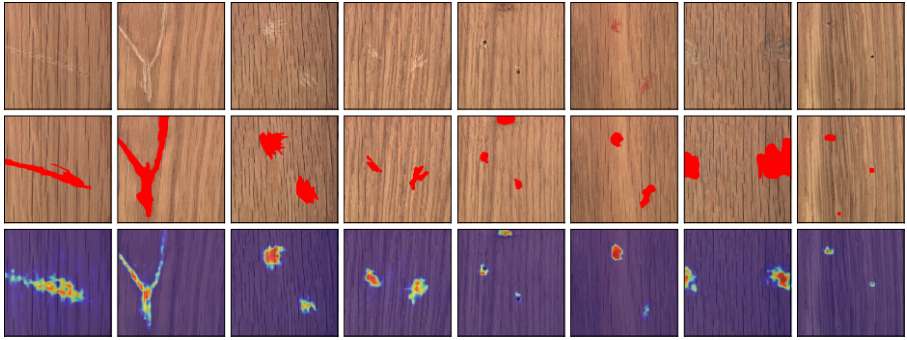

**Fig. 10: Visualization of anomaly maps generated by AdaCLIP for the wood category in MVTec AD.** The first row displays the input images, while the second row shows the ground truth. The bottom row illustrates the anomaly maps generated by AdaCLIP.

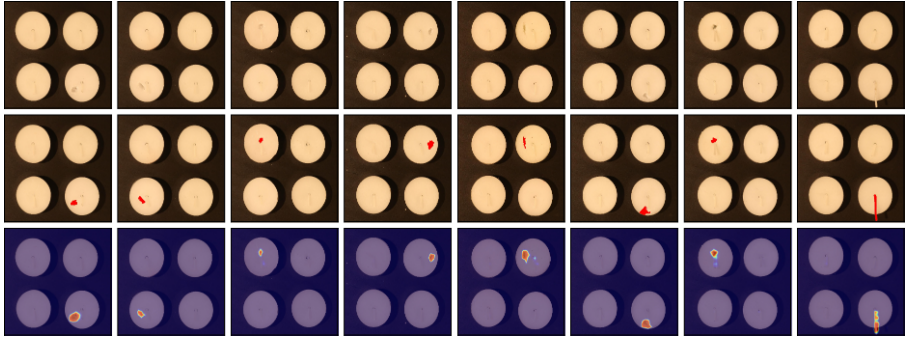

**Fig. 11: Visualization of anomaly maps generated by AdaCLIP for the candle category in VisA.** The first row displays the input images, while the second row shows the ground truth. The bottom row illustrates the anomaly maps generated by AdaCLIP.

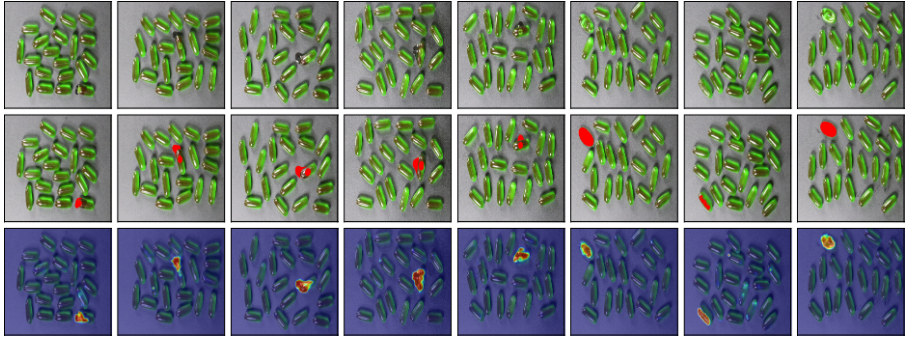

**Fig. 12: Visualization of anomaly maps generated by AdaCLIP for the capsules category in VisA.** The first row displays the input images, while the second row shows the ground truth. The bottom row illustrates the anomaly maps generated by AdaCLIP.

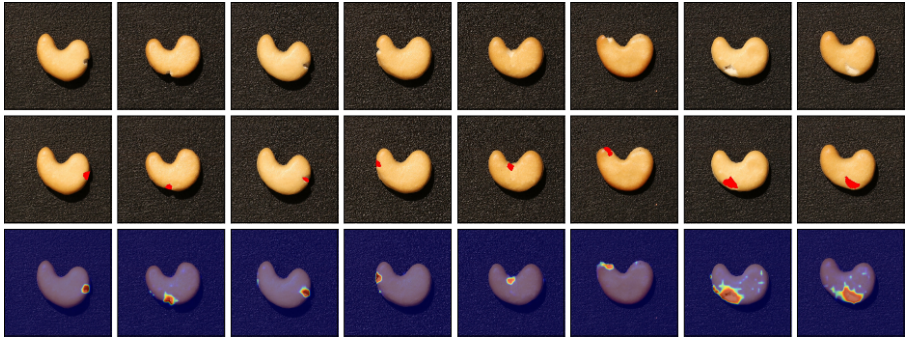

**Fig. 13: Visualization of anomaly maps generated by AdaCLIP for the cashew category in VisA.** The first row displays the input images, while the second row shows the ground truth. The bottom row illustrates the anomaly maps generated by AdaCLIP.

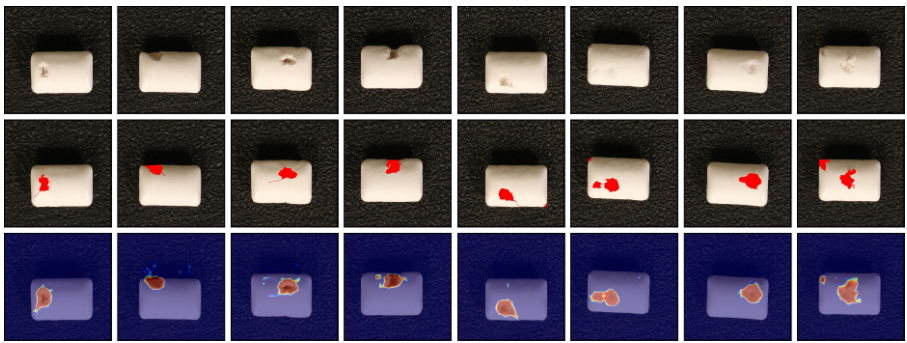

**Fig. 14: Visualization of anomaly maps generated by AdaCLIP for the chewinggum category in VisA.** The first row displays the input images, while the second row shows the ground truth. The bottom row illustrates the anomaly maps generated by AdaCLIP.

18

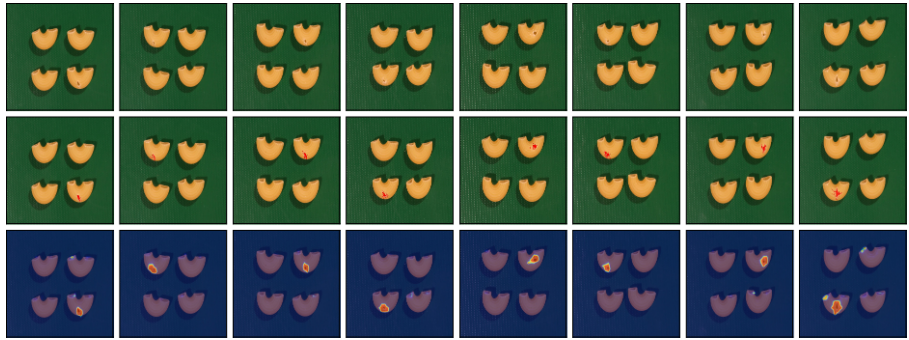

**Fig. 15: Visualization of anomaly maps generated by AdaCLIP for the macaroni1 category in VisA.** The first row displays the input images, while the second row shows the ground truth. The bottom row illustrates the anomaly maps generated by AdaCLIP.

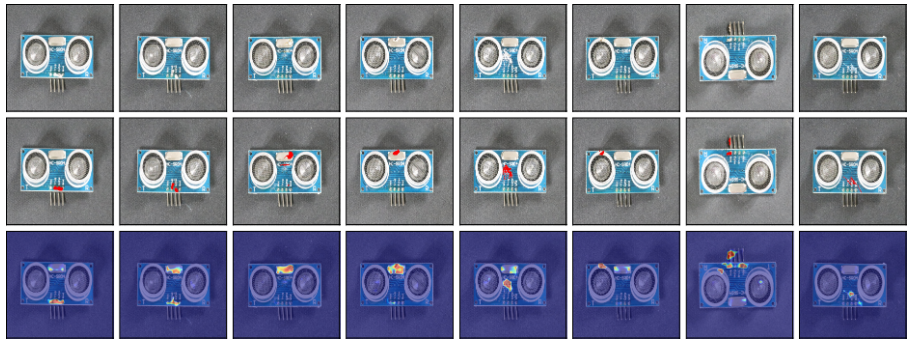

**Fig. 16: Visualization of anomaly maps generated by AdaCLIP for the pcb1 category in VisA.** The first row displays the input images, while the second row shows the ground truth. The bottom row illustrates the anomaly maps generated by AdaCLIP.

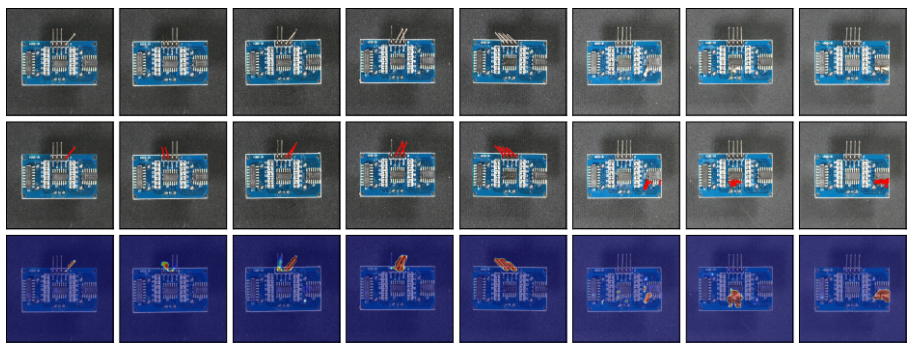

**Fig. 17: Visualization of anomaly maps generated by AdaCLIP for the pcb2 category in VisA.** The first row displays the input images, while the second row shows the ground truth. The bottom row illustrates the anomaly maps generated by AdaCLIP.

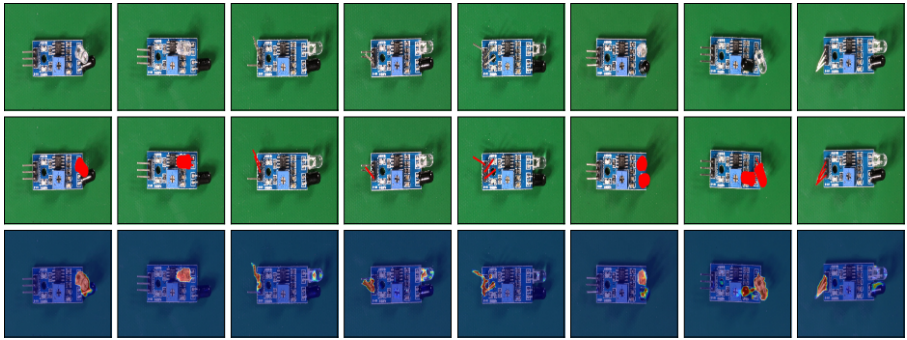

**Fig. 18: Visualization of anomaly maps generated by AdaCLIP for the pcb3 category in VisA.** The first row displays the input images, while the second row shows the ground truth. The bottom row illustrates the anomaly maps generated by AdaCLIP.

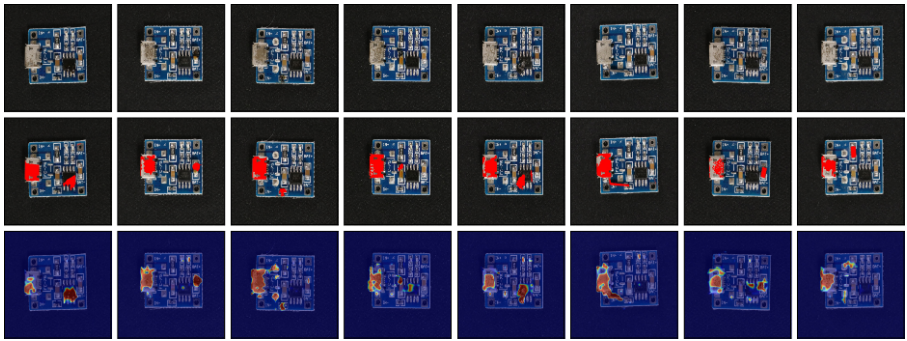

**Fig. 19: Visualization of anomaly maps generated by AdaCLIP for the pcb4 category in VisA.** The first row displays the input images, while the second row shows the ground truth. The bottom row illustrates the anomaly maps generated by AdaCLIP.

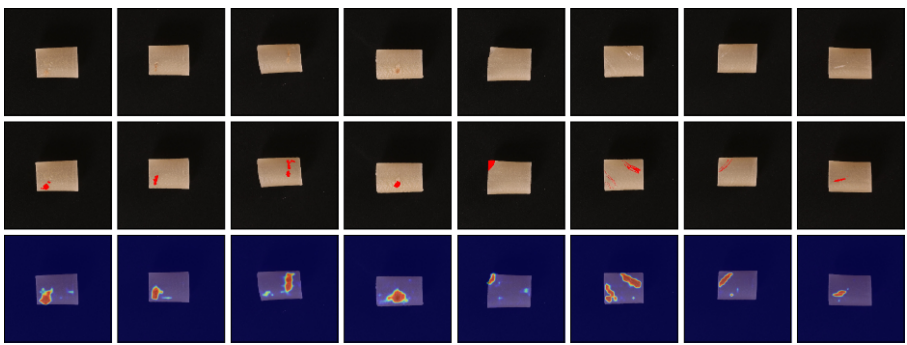

**Fig. 20: Visualization of anomaly maps generated by AdaCLIP for the pipe_fryum category in VisA.** The first row displays the input images, while the second row shows the ground truth. The bottom row illustrates the anomaly maps generated by AdaCLIP.

20

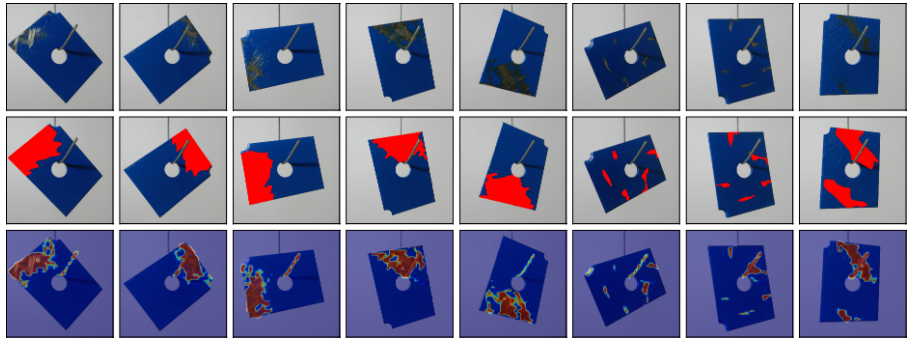

**Fig. 21: Visualization of anomaly maps generated by AdaCLIP for the metal plate category in MPDD.** The first row displays the input images, while the second row shows the ground truth. The bottom row illustrates the anomaly maps generated by AdaCLIP.

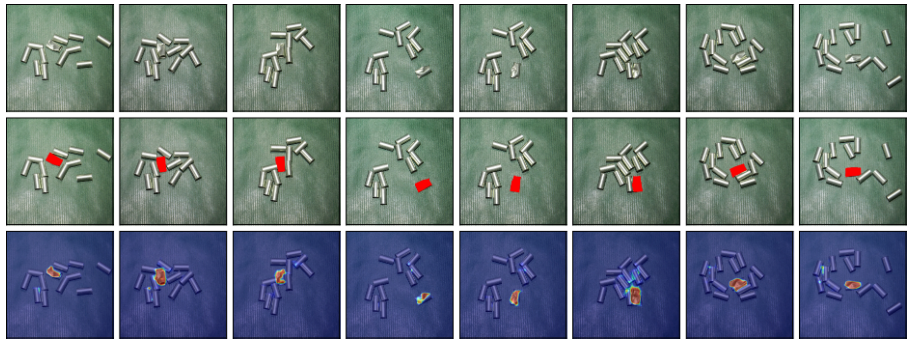

**Fig. 22: Visualization of anomaly maps generated by AdaCLIP for the tubes category in MPDD.** The first row displays the input images, while the second row shows the ground truth. The bottom row illustrates the anomaly maps generated by AdaCLIP.

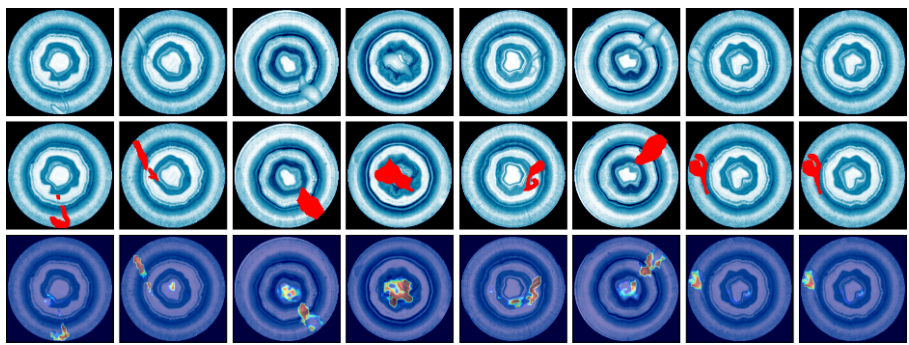

**Fig. 23: Visualization of anomaly maps generated by AdaCLIP for the class03 category in BTAD.** The first row displays the input images, while the second row shows the ground truth. The bottom row illustrates the anomaly maps generated by AdaCLIP.

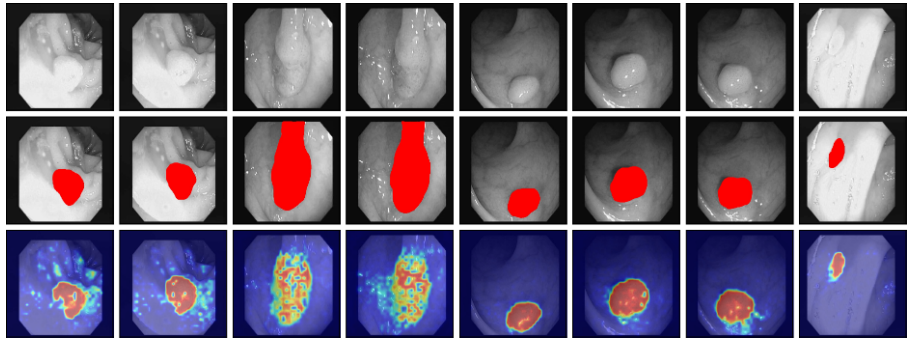

**Fig. 24: Visualization of anomaly maps generated by AdaCLIP for the Clinicdb dataset.** The first row displays the input images, while the second row shows the ground truth. The bottom row illustrates the anomaly maps generated by AdaCLIP.

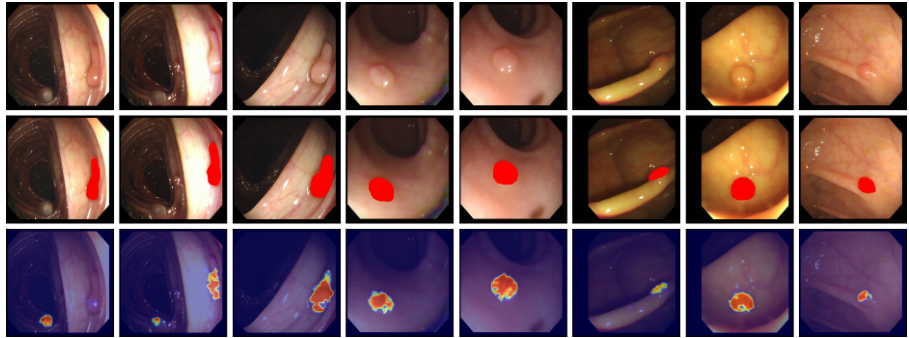

**Fig. 25: Visualization of anomaly maps generated by AdaCLIP for the Colondb dataset.** The first row displays the input images, while the second row shows the ground truth. The bottom row illustrates the anomaly maps generated by AdaCLIP.

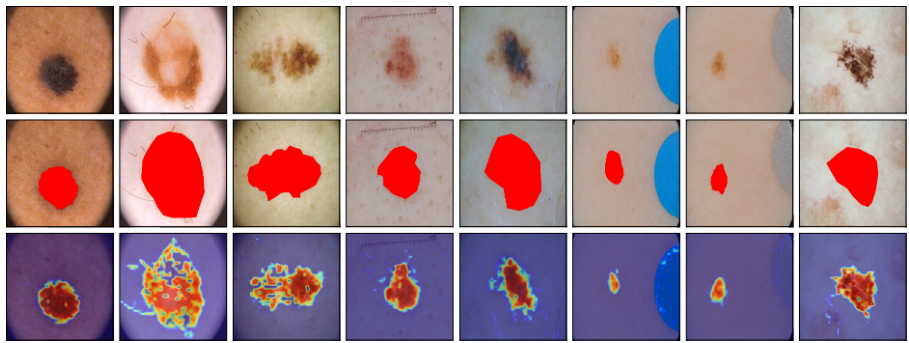

**Fig. 26: Visualization of anomaly maps generated by AdaCLIP for the ISIC dataset.** The first row displays the input images, while the second row shows the ground truth. The bottom row illustrates the anomaly maps generated by AdaCLIP.







\end{document}